\documentclass[conference]{IEEEtran}
\IEEEoverridecommandlockouts

\usepackage{cite}
\usepackage{amsmath,amssymb,amsfonts}
\usepackage{algorithmic}
\usepackage{graphicx}
\usepackage{textcomp}
\usepackage{xcolor}
\def\BibTeX{{\rm B\kern-.05em{\sc i\kern-.025em b}\kern-.08em
    T\kern-.1667em\lower.7ex\hbox{E}\kern-.125emX}}

\usepackage{amsmath}
\usepackage{amssymb}
\usepackage{booktabs}
\usepackage{mathtools}

\usepackage[utf8]{inputenc} %

\usepackage{calc}
\usepackage[font=small,skip=0pt]{caption}
\usepackage{circuitikz}
\usepackage{csquotes}
\usepackage{color}
\usepackage{colortbl}
\usepackage{xspace}

\usepackage[inline]{enumitem}
\usepackage[T1]{fontenc}
\usepackage[nolist, withpage, smaller]{acronym}
\usepackage{graphicx} %
\usepackage[unicode=true,
	hidelinks,
	pdfsubject={},
	pdfkeywords={},
	colorlinks=true,
	allcolors=blue,
	citecolor=blue
]{hyperref}

\usepackage{ifthen}

\usepackage{lipsum}

\usepackage{multirow}
\usepackage{multicol}

\usepackage{pgffor}
\usepackage{pgfgantt}
\usepackage{pgfplots, pgfplotstable}
\usepackage{rotating}
\usepackage{siunitx}
\usepackage{subcaption}
\usepackage{tabularx}
\usepackage{tikz}
\usepackage{tikzsymbols}
\usepackage{verbatim}
\usepackage{xargs}
\usepackage{xcolor}
\usepackage{colortbl}
\usepackage{xifthen}

\usepackage[capitalise]{cleveref}
\crefname{ALC@unique}{Line}{Lines}
\Crefname{ALC@unique}{Line}{Lines}

\usetikzlibrary{arrows}
\usetikzlibrary{arrows.meta}
\usetikzlibrary{automata}

\usetikzlibrary{backgrounds}
\usetikzlibrary{bending}

\usetikzlibrary{calc}
\usetikzlibrary{calendar}
\usetikzlibrary{chains}
\usetikzlibrary{circuits.logic.CDH}
\usetikzlibrary{circuits.logic.US}

\usetikzlibrary{decorations}
\usetikzlibrary{decorations.footprints}
\usetikzlibrary{decorations.fractals}
\usetikzlibrary{decorations.markings}
\usetikzlibrary{decorations.pathmorphing} %
\usetikzlibrary{decorations.pathreplacing}
\usetikzlibrary{decorations.shapes}
\usetikzlibrary{decorations.text}

\usetikzlibrary{er}

\usetikzlibrary{fadings}
\usetikzlibrary{fit}
\usetikzlibrary{folding}

\usetikzlibrary{matrix}
\usetikzlibrary{mindmap}

\usetikzlibrary{patterns}
\usetikzlibrary{petri}
\usetikzlibrary{pgfplots.colorbrewer} %
\usetikzlibrary{plotmarks}
\usetikzlibrary{positioning}

\usetikzlibrary{scopes}
\usetikzlibrary{shadows}
\usetikzlibrary{shapes}
\usetikzlibrary{shapes.arrows}
\usetikzlibrary{shapes.callouts}
\usetikzlibrary{shapes.gates.logic.US}
\usetikzlibrary{shapes.gates.logic.IEC}
\usetikzlibrary{shapes.geometric}
\usetikzlibrary{shapes.misc}
\usetikzlibrary{shapes.multipart}
\usetikzlibrary{shapes.symbols}

\usetikzlibrary{through}
\usetikzlibrary{tikzmark}
\usetikzlibrary{topaths}
\usetikzlibrary{trees}

\usepgfplotslibrary{colormaps}
\usepgfplotslibrary{colorbrewer} %
\usepgfplotslibrary{dateplot}
\usepgfplotslibrary{fillbetween}
\usepgfplotslibrary{patchplots}
\usepgfplotslibrary{groupplots}
\usepgfplotslibrary{statistics}

\pgfplotsset{compat=1.18}

\usepackage{algorithm}
\usepackage[colorinlistoftodos]{todonotes}

\usepackage{flushend}

\newif\ifdraft
	\draftfalse

	\definecolor{ColorHP}{RGB}{100,209,255}
	\definecolor{ColorNF}{RGB}{190,174,255}
	\definecolor{ColorGJ}{RGB}{204, 153, 0}
	\definecolor{ColorNS}{RGB}{0,153,76}
	\definecolor{ColorAC}{RGB}{255,51,51}

\newcommand{\cmpr}[1]{cf.~\cref{#1}}

\newcommand{\mycite}[1]{\cite{#1}}%

\newcommand{\plant}{f(x, u)}%

\newcommand{\decision}{\mathcal{D}}

\newcommand{\shift}{1cm}

\colorlet{ColorComplex}{white}
\colorlet{ColorDecision}{black!60}
\colorlet{ColorSimplex}{white}
\colorlet{ColorVoter}{ColorDecision}

\tikzset{arcnode/.style={rectangle, draw=black, rounded corners=0.125cm, minimum height=0.33*\shift, align=center, fill=white}}
\tikzset{switchhighlightnode/.style={draw=red!80!black!20!white, fill=red!80!black!20!white, line width=0.25mm, minimum width=8.5*\shift, minimum height=3.75*\shift, rounded corners}}
\tikzset{flowedge/.style={-latex, black, line width=0.25mm}}

\pgfdeclarelayer{farback}
\pgfsetlayers{farback,background,main}

\NewDocumentCommand{\Sensor}{ O{0} O{1.0} O{Sensor} O{black} m}{
	\begin{scope}[scale=#2]
		\node[anchor=south, draw, fill=#4!20, minimum width=0.2cm, minimum height=0.2cm, inner sep=0pt, scale=#2, rotate=#1] (SBox) at (#5) {};
		\node[minimum width=0.2cm, minimum height=0.4cm, inner sep=0pt, scale=#2, rotate=#1] (#3) at (SBox.north) {};

		\coordinate (A) at ($(SBox.north)+({-0.1*cos(-#1)+0.2*sin(-#1)},{0.1*sin(-#1)+0.2*cos(-#1)})$);
		\coordinate (B) at ($(SBox.north)+({0.1*cos(-#1)+0.2*sin(-#1)},{-0.1*sin(-#1)+0.2*cos(-#1)})$);
		\draw[fill=#4!40] (SBox.north) -- (A) -- (B) -- cycle;

	\end{scope}
}

\begin{document}

\begin{acronym}[ECU]

	\acro{ann}[ANN]{Artificial Neural Network}
	\acro{airgemm}[AIRGEMM]{AI and Robotics for GeoEnvironmental Modeling and Monitoring}
	\acro{ai}[AI]{Artificial Intelligence}
	\acro{app}[APP]{Adaptive Pure Pursuit}
	\acro{ar}[AR]{Augmented Reality}
	\acro{agv}[AGV]{Autonomous Ground Vehicle}
	\acro{asv}[ASV]{Autonomous Surface Vehicle}
	\acro{bc}[BC]{Behavioral Cloning}
	\acro{cave}[CAVE]{Cave Automatic Virtual Environment}
	\acro{cbf}[CBF]{Control Barrier Function}
	\acro{clf}[CLF]{Common Lyapunov Function}
	\acro{cps}[CPS]{Cyber-Physical System}
	\acro{dwa}[DWA]{Dynamic Window Approach}
	\acro{dwb}[DWB]{Dynamic Window Approach Version B}
	\acro{dm}[DM]{Decision Module}
	\acro{gps}[GPS]{Global Positioning System}
	\acro{gnss}[GNSS]{Global Navigation Satellite System}
	\acro{ivp}[IVP]{Initial Value Problem}
	\acro{lidar}[LiDAR]{Light Detection and Ranging}
	\acro{ml}[ML]{Machine Learning}
	\acro{mppi}[MPPI]{Model Predictive Path Integral}
	\acro{mpc}[MPC]{Model Predictive Control}
	\acro{nn}[NN]{Neural Network}

	\acro{odd}[ODD]{Operational Design Domain}
	\acro{ode}[ODE]{Ordinary Differential Equation}
	\acro{radar}[Radar]{Radio Detection and Ranging}
	\acro{rpp}[RPP]{Regulated Pure Pursuit}
	\acro{pp}[PP]{Pure Pursuit}

	\acro{hmd}[HMD]{Head-Mounted Display}
	\acro{hmi}[HMI]{Human Machine Interface}
	\acro{hp}[HP]{High-Performance}
	\acro{ha}[HA]{High-Assurance}

	\acro{rl}[RL]{Reinforcement Learning}
	\acro{roa}[RoA]{Region of Attraction}
	\acro{ros}[ROS]{Robot Operating System}
	\acro{ros2}[ROS2]{Robot Operating System Version 2}
    \acro{rta}[RTA]{Runtime Assurance}
	\acro{sonar}[Sonar]{Sound Navigation and Ranging}

	\acro{tcp}[TCP]{Transmission Control Protocol}
	\acro{teb}[TEB]{Timed Elastic Band}
	\acro{td3}[TD3]{Twin Delayed Deep Deterministic Policy Gradient}
	\acro{td3bc}[TD3+BC]{Twin Delayed Deep Deterministic Policy Gradient with Behavioral Cloning}
	\acro{tubaf}[TUBAF]{Freiberg University of Mining and Technology}
	\acro{uav}[UAV]{Unmanned Aerial Vehicle}

	\acro{vr}[VR]{Virtual Reality}
	\acro{xr}[XR]{Extended Reality}
\end{acronym}

\title{Context-dependent Runtime Construction of Safe Regions of Attraction from Design-Time Identification for Multi-Candidate Simplex Architectures}

\author{\IEEEauthorblockN{Georg Jäger, Nils-Jonathan Friedrich, Norman Seyffer}
\IEEEauthorblockA{\textit{Faculty of Mathematics and Computer Science} \\
\textit{TU Bergakademie Freiberg}\\
Freiberg, Germany \\
\{georg.jaeger, nils-jonathan.friedrich, norman.seyffer\}@informatik.tu-freiberg.de}
\and
\IEEEauthorblockN{José Cecílio, António Casimiro}
\IEEEauthorblockA{\textit{LASIGE, Faculty of Sciences} \\
\textit{University of Lisbon}\\
Lisbon, Portugal \\
\{jmcecilio, casim\}@ciencias.ulisboa.pt}
}

\maketitle

\begin{abstract}
	Deploying safety-critical cyber-physical systems requires control strategies that reconcile high performance and adaptability with rigorous safety guarantees. 
Classical control strategies offer stability proofs but are conservative, while advanced optimization- and learning-based controllers achieve high performance but lack formal verification. 
\ac{rta} architectures bridge this gap by introducing their own (dis-)advantages.
Safety filters use online optimization and introduce chattering, while Simplex permissiveness is bounded by the tightness of the certified \ac{roa} of its \ac{ha} controller. 
Targeting this conservatism, we present a two-stage \ac{roa} certification procedure leveraging contraction-based stability to establish a target set for Taylor-model-based reachability analysis. 
We then treat the closed-loop system under \ac{ha} control as a switched system, enabling: (a) runtime construction of context-dependent safe \acp{roa} and (b) safe switching between \ac{ha} candidates to optimize recovery and maximize the operating envelope of the uncertified \ac{hp} controller. 
We validate this framework on a robotic path-tracking task using a three-stage evaluation: \ac{ode} simulation for theoretical verification, high-fidelity Gazebo simulation under unmodeled dynamics, and experiments on physical hardware. 
Results show that the Simplex architecture maintains safety under fault injection and successfully shields a behavioral-cloning-trained \ac{hp} controller. 
Crucially, the decision module introduces negligible overhead (\SI{0.028}{\milli\second} average, \SI{4.13}{\milli\second} worst-case), consuming under \SI{10}{\percent} of the control period.

\end{abstract}


\acresetall
\section{Introduction}
\label{sec:introduction}

The deployment of autonomous \ac{cps} in safety-critical but complex and dynamic environments motivates a fundamental shift in control design.
It requires reconciling high performance and adaptability with strict safety guarantees.
Commonly, safety-critical systems rely on formal proofs to guarantee safety, which are conservative and limit capability~\mycite{papachristodoulou2002construction, chang2019neural}.
Conversely, advanced optimization- or learning-based controllers excel in dynamic environments but lack safety certificates due to their complex, black-box nature.

\ac{rta} addresses this conflict by decoupling performance from safety.
These architectures monitor the system state and the actions proposed by the uncertified \ac{hp} controller.
If deemed unsafe, inputs are overwritten, e.g., through safety filters or blending architectures, or control is switched to a verified \ac{ha} controller to maintain a safe operating envelope as done in the Simplex architecture~\mycite{sha2001using}.
It permits \ac{hp} control as long as the \ac{ha} controller is guaranteed to recover the system.
A \ac{dm} monitors this condition, switching to the \ac{ha} controller whenever continued \ac{hp} execution would risk leaving the \ac{roa}. %

However, the practical value of this architecture depends on the conservatism of the \ac{ha}'s approximated safe \ac{roa}.
While over-approximation is unsafe, under-approximation forces the \ac{dm} to switch prematurely, degrading overall system performance.
Existing methods address this partially.
Lyapunov and contraction analyses~\mycite{forni2013differential} guarantee formal safety but yield conservative inner approximations.
Reachability analysis~\mycite{dongxu2020reachability} extends coverage but suffers from the wrapping effect.
Furthermore, using a single \ac{ha} controller neglects the complementary coverage advantages a set of candidates can provide.

We leverage these observations for our contributions.
First, we present a two-stage procedure for certifying the \ac{roa} of a controller. 
A contraction-based stability analysis establishes a core \ac{roa} that is used by Taylor-model reachability~\mycite{li2025verifying} as a target set and incrementally extended.
In both stages, we use partitioning of the parameter and state spaces to limit wrapping effects and further reduce conservatism, thereby generating a certified set of safe \acp{roa} for a controller.
Along with other design-time artifacts, e.g., the maximal dwell time, we refer to this as the controller's \textit{recovery profile} $\mathcal{R}$.
This is used at runtime to construct the context-dependent \ac{roa}, which we leverage in our second contribution.
There, we treat the closed-loop system under the \ac{ha} controller as a switched system over a set of \ac{ha} candidates. 
This lets the \ac{dm} not only switch between \ac{hp} and \ac{ha} modes, but also between \ac{ha} candidates to optimize, e.g., recovery time.

To motivate the use of the Simplex architecture, \cref{sec:soa_shielding} reviews the state of the art on \ac{rta} architectures, where we further analyze safety filters and control blending as alternatives before considering approaches for identifying \acp{roa}.
\Cref{sec:concept} then derives and discusses our main contributions.
Although use case agnostic, we use path tracking of a differential-drive mobile robot in outdoor scenarios (e.g., urban parcel delivery), with \ac{rpp}~\mycite{macenski2023regulated} in the \ac{ha} role as a running example to support explanations.
The resulting Simplex controller is evaluated in \cref{sec:evaluation} in three stages: \ac{ode}-based simulation to validate the switching logic under the modeled dynamics, Gazebo simulation to analyze the prediction horizon needed under realistic dynamics, and real-world experiments on the platform shielding an \ac{ml} controller based on behavioral cloning.
Finally, \cref{sec:conclusions} concludes and outlines directions for future work.

\section{Runtime Assurance of High-Performance Controllers}
\label{sec:soa_shielding}

For \ac{rta} of systems against unsafe actions of unverified \ac{hp} controllers, multiple approaches exist.
We evaluate these approaches across four common criteria: (i) the strength of formal safety guarantees, (ii) the conservatism imposed on performance, (iii) computational cost for real-time execution, and (iv) the required analytical certification effort.
Accordingly, we review \ac{cbf}-based safety filters (\cmpr{subsec:soa_cbfs}) and identify their chattering problem as well as computational overhead caused by online optimization.
We further review blending architectures (\cmpr{subsec:soa_blending}), which are challenging to verify due to their continuous handover of control, before evaluating reachability-based \ac{rta} (\cmpr{subsec:soa_rta}), which similarly causes computational overhead and suffers from the wrapping effect.
Given these challenges, we position the Simplex architecture (see \cmpr{subsec:simplex_architecture}) as a suitable alternative for resource-constrained, time-sensitive applications.
Finally, we discuss methods for estimating a controller's \ac{roa} (\cmpr{subsec:soa_roa}), establishing the foundation for \cref{sec:concept}.

\subsection{Safety Filters based on Control Barrier Functions}
\label{subsec:soa_cbfs}

Safety filters use \acp{cbf} similar to cost functions for enforcing safety constraints on actions proposed by an unverified controller~\mycite{wang2025safe}.
Once the monitored action is deemed unsafe, online optimization is used to minimize the alteration of the proposed action.
Through monitoring and, if needed, adjusting the control action in every control period, safety filters can guarantee forward invariance of the safe set~\mycite{wang2025safe, almubarak2022safety}. 

However, in contrast to convex quadratic programs applicable to control-affine dynamics, online optimization for non-control-affine systems requires non-convex nonlinear programs where satisfying deterministic worst-case execution time (WCET) bounds is difficult. 
Furthermore, these myopic formulations risk recursive infeasibility under input bounds, conflicting constraints, or robust margins, compromising safety guarantees upon solver failure or delay~\mycite{garg2024advances}.
Another aspect is the effort required to construct a valid \ac{cbf} for a nonlinear system with input constraints.
It is an open problem that has recently been addressed with \ac{ml} approaches as well, e.g.,~\mycite{10491341}.
Furthermore, near the safe set boundaries, small state fluctuations can cause the safety constraints to repeatedly toggle on and off.
This induces high-frequency control chattering, non-smooth control actions, degrades performance, and causes excessive actuator wear~\mycite{brunke2025preventing}.

\subsection{Blending Architectures}
\label{subsec:soa_blending}

Instead of overwriting the control input, \textit{blending} architectures aim at a continuous handover from \ac{hp} control to \ac{ha} through a state-dependent weighting function $\sigma(x) \in [0, 1]$.
This yields a shared control policy \mbox{$u(x) = (1 - \sigma(x))\, C_{HP}(x) + \sigma(x)\, C_{HA}(x)$}~\mycite{gohari2022blending}.
As the system approaches a safety boundary, $\sigma(x)$ smoothly scales up the influence of the \ac{ha} controller, producing more continuous actuator commands than hard switching.

However, designing $\sigma(x)$ such that the closed-loop system can be analytically proven safe and stable is demanding, increasing the certification effort.
Because the dynamics are driven by a convex combination of two distinct vector fields, the closed-loop behavior can exhibit unintended equilibria or mutual cancellation of conflicting control efforts~\mycite{shen2024combining}.
Consequently, the formal safety guarantees can induce performance conservatism. 
Ensuring correctness across the full state space may force $\sigma(x)$ to engage the \ac{ha} controller before it is strictly necessary, degrading the performance that the \ac{hp} controller was introduced to provide.
On the other hand, the computational cost of computing context-aware blending weights in real time can be adjusted according to safety requirements and available resources, depending on the use case at hand. 

\subsection{Reachability-Based Runtime Assurance}
\label{subsec:soa_rta}

Reachability-based \ac{rta} frameworks ensure safety by predicting the future state evolution of an uncertified \ac{hp} controller using online set propagation~\mycite{bak2014real, musau2022using, althoff2021set, dongxu2020reachability}.
Starting from either a single state or a set of states, an \ac{rta} module computes the forward reachable set (a sequence of sets propagated by the system dynamics) of the system over a predictive horizon.
If the intersection of the predicted reach-tube with any hazardous state becomes non-empty, or if the prediction indicates that the system state risks leaving a recoverable backup envelope, the \ac{rta} framework overrides the \ac{hp} controller and engages a verified fallback control.

The resulting formal safety guarantee is established by an explicit per-trajectory proof at each control cycle, thereby reducing conservatism as currently irrelevant regions of the parameter space are ignored by considering only the active contextual situation.
However, shifting certification to runtime introduces significant computational and performance challenges.
First, regardless of the geometric representation, set propagation suffers from the wrapping effect~\mycite{li2025verifying, althoff2021set}.
Efficient convex representations (intervals, ellipsoids, polytopes, zonotopes~\mycite{luo2023reachability}) struggle under nonlinear dynamics, where bounding the curved, non-convex reachable sets at each step introduces accumulating over-approximation (the nonlinear wrapping effect).
To mitigate this, non-convex representations propagate state dependencies alongside the set. 
Although polynomial zonotopes~\mycite{kochdumper2020sparse} provide exact linear mapping, they suffer from combinatorial generator growth under general dynamics. 
Taylor models~\mycite{chen2012taylor} resolve this by combining a high-degree multivariate polynomial (capturing the curved shape) with an interval remainder (bounding truncation errors). 
By approximating vector fields directly through Taylor series expansions, this approach is applicable to arbitrary non-polynomial nonlinearities (e.g., trigonometric kinematics) to suppress the wrapping effect over extended horizons. 
The method's accuracy and complexity are controlled by the time order $orderT$ (governing time integration) and jet transport order $orderS$ (limiting the polynomial degree). 
Nevertheless, repeatedly over-approximating sets to limit complexity still accumulates errors over time.
This performance conservatism causes premature fallback interventions, restricting the operational domain and limiting the \ac{hp} controller's utility~\mycite{miller2024optimal}.

Second, the real-time computational cost of computing nonlinear forward reach-tubes dynamically is high. 
Tight reachability methods for nonlinear systems are typically too intensive for online computation and are therefore commonly performed offline~\mycite{herbert2019reachability, everett2021efficient}.

On embedded platforms subject to strict processor budgets and tight control rates, guaranteeing that reachability calculations complete within deterministic worst-case execution-time bounds is a major challenge. 
If the online prediction fails to complete within a strict sampling period, runtime safety verification is compromised~\mycite{bak2014real}.

\subsection{Simplex Architecture and Switching Logic}
\label{subsec:simplex_architecture}
\label{subsec:switching_logic}

Given the chattering of safety filters, the difficulty of formally certifying blending architectures, and the computational overhead of reachability-based \ac{rta}, the Simplex architecture~\mycite{sha2001using} constitutes an alternative suited for resource-constrained real-time systems.
It consists of three main components: a \ac{hp} controller $C_{HP}(x)$, a \ac{ha} controller $C_{HA}(x)$, and a decision module $\decision$ (with $x \in \mathcal{X} \subseteq \mathbb{R}^n$ being the state of the system). %
As before, a \ac{hp} controller proposes an action, which is verified by the decision module.
By deciding whether to switch control from the \ac{hp} to the \ac{ha} controller if deemed unsafe, and back once the system recovers, the decision module plays a critical role in the system.
Thus, the success of the architecture's paradigm depends on the derivation of a safe but performant switching logic.

Approaches to deriving the same trade-off between real-time cost and performance conservatism can be grouped into three categories (Analytical approximations, Proofs-on-demand, and Static set approximations).

\textit{Analytical approximations} precompute algebraic switching conditions to minimize runtime computation. 
Taylor expansions of Barrier Certificates with remainder bounds~\mycite{damare2025barrier} or Forward Switching Conditions~\mycite{mehmood2023distributed} derived from \acp{cbf} rely on predicting only one control period to define switching conditions.
Such checks incur minimal runtime cost, but their guarantees depend on the accuracy of the underlying certificates and grow conservative when these are restrictive.

\textit{Proofs-on-demand} architectures shift certification to runtime. 
They extend the online reachability to the recoverable region ad-hoc when a one-step prediction leaves it~\mycite{maderbacher2025adaptive}, while black-box variants verify candidate command sequences entirely at runtime, falling back to stored safe profiles upon timeout~\mycite{sheikhi2024blackbox}.
This maximizes permissiveness but, risks timing violations on embedded platforms, \cmpr{subsec:soa_rta}.

\textit{Static set approximations} occupy the middle ground by approximating the \ac{roa} as a fixed maximum-volume Lyapunov ellipsoid~\mycite{nesti2026simplex, salamini2025poster}, which enables low-cost membership checks but remains conservative for nonlinear systems.

To reduce conservatism, multi-layer resilient architectures employ compound \ac{ha} subsystems rather than a single monolithic fallback controller~\mycite{shi2026multilayer}. 
\mycite{shi2026multilayer} studies the use case of a \ac{uav}, but does not provide proofs for safety and stability under the layered control regime.

In conclusion, deriving a safe switching logic is constrained by how the \ac{ha} controller's recoverable region is characterized.
Current methods either rely on online reachability to limit conservatism but incur computational overhead, or use static set approximations whose low overhead on resource-constrained devices is offset by performance conservatism.
To address this trade-off, the next subsection reviews techniques for estimating these \acp{roa} at design time.

\subsection{Estimating Regions of Attraction}
\label{subsec:soa_roa}

Reducing runtime computational overhead requires analyzing the \ac{ha} controller at design time to generate artifacts for defining non-conservative switching rules. 
A central characteristic in that respect is a controller's \ac{roa}: the set of initial states $\mathcal{A} \subseteq \mathcal{X}$ from which the closed-loop system is guaranteed to stabilize.
Here, we analyze common Lyapunov stability, contraction analysis, and revisit reachability analysis, which require increasing computational effort for reducing conservatism.

\subsubsection{Lyapunov Analysis}
\label{subsubsec:lyapunov_roa}

Lyapunov stability is established by showing that a positive-definite function $V(x)$ strictly decreases along system trajectories ($\dot{V}(x) < 0$ for $x \neq 0$).
For a linear system $\dot{x} = Ax$, we can assume a quadratic function $V(x) = x^T P x$ and solve the continuous algebraic Lyapunov equation $A^T P + P A = -Q$ for $P = P^T \succ 0$ with $Q \succ 0$ to validate this condition. 
The sublevel set $\mathcal{E} = \{x \mid x^T P x \le c\}$ then serves as a forward-invariant estimate of the \ac{roa}.
For nonlinear systems, standard extensions include Sum-of-Squares (SOS) programming for polynomial functions~\mycite{papachristodoulou2002construction} and neural Lyapunov methods parameterized by deep neural networks~\mycite{chang2019neural}.
However, these techniques nevertheless entail conservatism and require complex verification pipelines.
Moreover, requiring a single common Lyapunov function to satisfy descent conditions across all parameter variations forces conservative worst-case approximations that under-estimate the parameter-dependent \acp{roa}.
To mitigate this, switched systems theory uses separate, mode-specific Lyapunov functions for individual parameter regimes~\mycite{liberzon1999basic, decarlo2000perspectives}.
Stability under switching is then guaranteed by either proving that the active energy decreases at each switch, or enforcing a minimum dwell time between transitions to allow energy dissipation~\mycite{liberzon1999basic, cao2010dwell}.

\subsubsection{Contraction Analysis}
\label{subsubsec:contraction_roa}

Contraction analysis~\mycite{lohmiller1998contraction} generalizes standard Lyapunov stability to incremental stability, assessing whether neighboring system trajectories displaced by an infinitesimal virtual displacement $\delta x$ converge toward one another~\mycite{forni2013differential}.
Specifically, $\delta x$ evolves according to the dynamics $\dot{\delta x} = J(x) \delta x$, where $J(x)$ is the system Jacobian at state $x$.
By defining a coordinate-transformed displacement $\delta z = \Theta \delta x$ under a constant transformation matrix $\Theta^T \Theta = P$ to allow less conservative non-Euclidean metrics, the transformed displacement dynamics are given by:
\begin{equation}
    \dot{\delta z} = \bar{J} \delta z, \quad \text{with } \bar{J} = \Theta J(x) \Theta^{-1}
    \label{eq:transformed_variational}
\end{equation}
The rate of change of the infinitesimal distance squared $\|\delta z\|^2 = \delta x^T P \delta x$ is then given by $\frac{d}{dt} \|\delta z\|^2 = 2 \delta z^T \bar{J}_s \delta z$.
Therefore, any virtual displacement contracts if the symmetric part of the coordinate-transformed Jacobian is uniformly negative definite:
\begin{equation}
    \bar{J}_s = \frac{1}{2}(\bar{J} + \bar{J}^T) \prec 0
    \label{eq:contraction_jacobian}
\end{equation} 
Note that this is equivalent to requiring the matrix $Q$ to be positive-definite in standard linear Lyapunov analysis.
Moreover, under a constant metric $P$, contraction analysis can be interpreted as applying Lyapunov theory differentially. 
Certifying that the transformed Jacobian is negative definite ($\bar{J}_s \prec 0$) pointwise guarantees that the standard quadratic Lyapunov function $V(x) = x^T P x$ strictly decreases, ensuring that the sublevel set $\mathcal{E} = \{x \mid x^T P x \le c\}$ remains forward-invariant.
Unlike standard Lyapunov analysis, this contraction condition can be verified over intervals of the state and parameter space.
However, both approaches remain restricted to estimating symmetric, convex stability regions whose shape is dictated by $P$, which may fail to capture non-convex, potentially asymmetric boundaries of the \ac{roa}.

\subsubsection{Reachability Analysis}
\label{subsubsec:reachability}

While we have already discussed reachability analysis for online \ac{rta}, applying it at design time can overcome conservatism and the convexity restrictions of Lyapunov-based methods.
For instance, estimating the \ac{roa} via forward reachable sets bypasses the need for a Lyapunov function, yielding accurate, non-convex estimates~\mycite{elguindy2017estimating}.
Similarly, precomputing sequences of one-step contractive terminal sets offline enlarges the certified region without increasing online computational overhead~\mycite{limon2005enlarging}.
Furthermore, explicit and implicit reachability formulations -- including backward reachable sets from Hamilton-Jacobi-Isaacs (HJI) equations -- map the exact geometry of the stability boundary, significantly reducing conservatism compared to traditional energy-based certificates~\mycite{yuan2023comparing}.

\section{Multi-Candidate Simplex Architecture}
\label{sec:concept}

\begin{figure*}[!t]
    \centering
    \begin{minipage}[t]{0.2\textwidth}
        \centering
        \resizebox{\columnwidth}{!}{\input{figures/generated/roa/configuration-search/C4/traj_id0658_kappa_0.90_safetrue.pgf}}
        \caption{Reach-tube simulation demonstrating recovery.}
        \label{fig:ha_sim}
    \end{minipage}\hfill
    \begin{minipage}[t]{0.39\textwidth}
        \centering
        \resizebox{\columnwidth}{!}{\input{figures/generated/roa/full-kappa/C4/roa_intersection_bin1_bin20.pgf}}
        \caption{Safe \acp{roa} for two opposite curvature bins and their intersection.}
        \label{fig:roa_intersection}
    \end{minipage}\hfill
    \begin{minipage}[t]{0.39\textwidth}
        \centering
        \resizebox{\columnwidth}{!}{\begin{tikzpicture}[circuit logic CDH]
    \node[arcnode, minimum width=1.5cm] (p) at (1.25,-0.95) {Plant $\plant$};
    \node[or gate US, draw,logic gate inputs=pp, text width=1.5cm, align=center, fill=white, minimum height=2cm] (d) at (3.25,0.5) {Decision $\decision$};

    \node[arcnode, minimum width=1.75cm] (c) at ($(d.input 1) + (-2.25,0)$) {High-Perf. $C_{HP}(x)$};
    
    \node[arcnode, minimum width=1.75cm] (s3) at ($(d.input 2) + (-2.25,0) + (0.15,-0.35)$) {High-Ass. $C_{HA,M}(x)$};
    \node[arcnode, minimum width=1.75cm] (s2) at ($(d.input 2) + (-2.25,0) + (0.075,-0.175)$) {High-Ass. $C_{HA,2}(x)$};
    \node[arcnode, minimum width=1.75cm] (s) at ($(d.input 2) + (-2.25,0)$) {High-Ass. $C_{HA,1}(x)$};

    \Sensor[90][2.5][Sensor]{$(c.west)!0.5!(s.west) + (-0.25*\shift,0)$}
    \node[anchor=north] (SensorLabel) at ($(Sensor.west) + (0.0, 0.05*\shift)$){Sensor};

    \draw[flowedge] (c.east) -- node[above] {$u_{HP}$} (d.input 1);
    
    \begin{pgfonlayer}{background}
        \draw[flowedge] (s3.east) -- ($(d.input 2) + (0,-0.35)$);
        \draw[flowedge] (s2.east) -- ($(d.input 2) + (0,-0.175)$);
    \end{pgfonlayer}
    \draw[flowedge] (s.east) -- node[above] {$u_{HA}$} (d.input 2);
    
    \draw[flowedge] (d.output) |- node[above, pos=0.8] {$u$} (p.east);
    \draw[flowedge] (p.west) -| node[above, pos=0.2] {$y$} ($(Sensor.west) + (0,-0.5)$) -- ($(Sensor.west)+(0,-0.35)$);
    \draw[flowedge] (Sensor.south) -- ([xshift=0.25cm]Sensor.south) |- node[above,pos=0.75] {$x$} (c.west);
    \draw[flowedge] (Sensor.south) -- ([xshift=0.25cm]Sensor.south) |- node[above,pos=0.75] {$x$} (s.west);
\end{tikzpicture}}
        \caption{The Simplex architecture using a family of \ac{ha} candidates.}
        \label{fig:simplex_architecture_concept}
    \end{minipage}
\end{figure*}

The last section underlined the trade-off between computational overhead during \ac{rta} that reduces conservatism by focusing on context- and parameter-dependent \ac{roa} and conservatism of static, design-time-based \ac{roa} approximations that minimize computational overhead.
In this section, we aim at bridging this gap by proposing a multi-candidate Simplex architecture whose switching rules construct context-dependent safe \acp{roa} at runtime from \ac{roa} results identified at design time.
The approach is generic, but we present it alongside a path-tracking use case for a differential-drive mobile robot for clarity.
After defining the use-case dynamics (\cref{subsec:system_definition}), we combine contraction analysis with Taylor-model forward reachability to reduce the conservatism of the identified \ac{roa}, \cmpr{subsec:roa}.
Partitioning the parameter space manages the wrapping effect of interval arithmetic and entails a switched system that enables context-dependent \ac{roa} approximations at runtime.
Then, we exploit this switched-system view in \cref{subsec:decision_module}, where the decision module's switching rules manage transitions between \ac{hp} and \ac{ha} modes, and between \ac{ha} candidates.

\subsection{System Definition}
\label{subsec:system_definition}

As an illustrative example, we consider path tracking for a differential-drive mobile robot with nonlinear kinematics:
\begin{equation}
	\dot{x} = v \cos(\theta), \quad \dot{y} = v \sin(\theta), \quad \dot{\theta} = \omega
	\label{eq:kinematics}
\end{equation}
where $(x,y) \in \mathbb{R}^2$ is the 2D position, $\theta$ is the orientation, and control inputs $u = [v, \omega]^T$ are issued by a controller $C(x)$.
The goal is to track a path $\mathcal{P} = \{p_0, \ldots, p_m\}$ of waypoints.
To analyze the \ac{roa}, we abstract $\mathcal{P}$ to a continuous reference path $\mathcal{P}_{\text{ref}}$ with local curvature $\kappa_{\text{ref}}$ and reformulate \cref{eq:kinematics} in the Frenet frame~\cite{werling2010optimal}, yielding the error dynamics
\begin{equation}
	\dot{e}_y = v \sin(e_\theta), \quad
	\dot{e}_\theta = \omega - \frac{v \cos(e_\theta)\, \kappa_{\text{ref}}}{1 - \kappa_{\text{ref}}\, e_y}
	\label{eq:error_dynamics}
\end{equation}
where $e_y$ is the signed lateral error, $e_\theta$ is the heading error, and the error state vector is $x = [e_y, e_\theta]^T \in \mathcal{X} = \mathbb{R} \times [-\pi, \pi]$.
The safety set $\mathcal{X}_\mathcal{S} \subseteq \mathcal{X}$ defines the safe error states.
For theoretical development, we assume a \ac{ha} controller $C_{HA}(x)$ safely stabilizing the dynamics is given. 
We define the concrete controller for evaluation in \cref{subsec:ha_search}.

\subsection{Runtime Construction of Context-Dependent \acp{roa} from Designtime Identification Results}
\label{subsec:roa}

Designing the decision module's switching rules requires characterizing the stability and safety regions of the \ac{ha} controller.
To minimize runtime overhead while reducing conservatism, we split this process: at design time, we use a two-stage pipeline to identify safe \acp{roa} (\cref{subsubsec:roa_stage_1,subsubsec:roa_stage_2}); at runtime, we dynamically construct the context-dependent safe \ac{roa} (\cref{subsubsec:roa_runtime}).
We consider the generic closed-loop dynamics $\dot{x} = f(x, C_{HA}(x), \lambda)$ with parameter space $\lambda \in \Lambda \subseteq \mathbb{R}^m$ representing uncertain or variable operational parameters (e.g., environmental conditions, physical properties, or the one-dimensional curvature range $\Lambda = [\underline{\kappa_{\text{ref}}}, \overline{\kappa_{\text{ref}}}]$ with the active parameter $\lambda = \kappa_{\text{ref}}$ in case of the path-tracking example).
For this system, the safe \ac{roa} $\mathcal{A}$ is the set of initial states $x_0$ whose trajectories remain within the safety set $\mathcal{X}_\mathcal{S}$ and asymptotically converge to a target set $\mathcal{E}$ under all parameter variations:
\begin{equation}
	\begin{split}
		\mathcal{A} = \{ x_0 \in \mathcal{X} \mid \, & \forall \lambda \in \Lambda, \, x(0)=x_0 \implies \\
		& \forall t \ge 0, \, x(t) \in \mathcal{X}_\mathcal{S} \land \lim_{t\to\infty} x(t) \in \mathcal{E} \}
	\end{split}
\end{equation}

\subsubsection{Stage 1 -- Contraction ellipsoid}
\label{subsubsec:roa_stage_1}

\begin{algorithm}[t]
\caption{Contraction-Based RoA Approximation}
\label{alg:contraction_analysis}
\footnotesize
\begin{algorithmic}[1]
\REQUIRE \ac{ha} controller $C_{HA}$, closed-loop dynamics $f(x, C_{HA}(x), \lambda)$, parameter space partition $\Lambda_P$, safety set $\mathcal{X}_\mathcal{S}$
\ENSURE Metric $P$, sublevel value $c$
\STATE \colorbox{gray!20}{$\text{Candidates} \leftarrow \{P \succ 0 \mid A^T P + P A = -I \text{ at } \lambda_j \in \Lambda \}$} \label{alglin:contraction_search}
\STATE $(P_{\text{best}}, c_{\text{best}}) \leftarrow (\text{None}, 0)$
\FORALL{$P \in \text{Candidates}$}
    \STATE \colorbox{gray!20}{\parbox{\dimexpr\linewidth-2\fboxsep\relax}{\raggedright $c \leftarrow \text{BinarySearchMax}(c \in [0, c_{\max}(P, \mathcal{X}_\mathcal{S})])$ s.t.\ \textsc{VerifyContraction}$(P, c, \Lambda_P, C_{HA})$}} \label{alglin:contraction_verify}
    \IF{$c > c_{\text{best}}$}
        \STATE $(P_{\text{best}}, c_{\text{best}}) \leftarrow (P, c)$
    \ENDIF
\ENDFOR
\STATE \RETURN $(P_{\text{best}}, c_{\text{best}})$
\STATE
\STATE \textbf{procedure} \textsc{VerifyContraction}($P, c, \Lambda_P, C_{HA}$)
\STATE \quad $\Theta \leftarrow \text{Cholesky}(P)$
\STATE \quad \textbf{verify} $\bar{\mathbf{J}}_s(B, \Lambda_k) \prec 0$ on $\mathcal{E} = \{x \mid x^T P x \le c\}$ for all $\Lambda_k \in \Lambda_P$
\STATE \quad via bisection of $B = \text{BoundingBox}(\mathcal{E})$ up to depth $d_{\max}$
\end{algorithmic}

\end{algorithm}

In the first stage, we aim to establish a common target set for forward reachability used in Stage~2.
We require a local region of asymptotic stability, $\mathcal{E}$, valid across the entire parameter envelope, $\Lambda$.
Since uncertain parameters possibly contained in $\Lambda$ require set-based propagation, we follow \cref{subsubsec:contraction_roa} and apply contraction theory.
To mitigate the entailed wrapping effect, we adopt a search-and-verify procedure in \cref{alg:contraction_analysis}, decoupling the generation of metric candidates from their formal verification.
The \textit{search phase} (\cref{alglin:contraction_search}) heuristically generates candidate metrics $P$ by sampling parameter points $\lambda_j \in \Lambda$ and solving the Lyapunov equation (\cref{subsubsec:lyapunov_roa}) with $Q = I$.
The resulting $P$ serves as the shape of an energy-like Lyapunov function of the linearized system at the equilibrium and as a valid distance metric for contraction analysis.
The latter perspective is leveraged in the \textit{verification phase} (\cref{alglin:contraction_verify}) to find the largest invariant ellipsoid $\mathcal{E} = \{x \mid x^T P x \leq c\}$ via binary search over $c$ for which contraction can be verified.
Verification requires the transformed Jacobian to be negative definite ($\bar{J}_s \prec 0$), \cmpr{subsubsec:contraction_roa}.
A single interval evaluation over the envelope $\Lambda$ would be overly conservative, so we partition the parameter space into hyperrectangles (bins) $\Lambda_P = \{\Lambda_k\}_{k=1}^K$ and verify negative definiteness on each sub-box via the coordinate-transformed interval Jacobian
\begin{equation}
	\bar{\mathbf{J}}_s(B, \Lambda_k) = \text{sym}\!\left(\Theta\, \mathbf{J}_f(B, \Lambda_k)\, \Theta^{-1}\right),
\end{equation}
with Cholesky factor $\Theta$ ($\Theta^T\Theta = P$), a state box $B \subseteq \mathcal{E}$, parameter bin $\Lambda_k \in \Lambda_P$, the symmetric matrix operator $\text{sym}(M) = \frac{1}{2}(M+M^T)$, and $\mathbf{J}_f(B, \Lambda_k)$ denoting the interval evaluation of the Jacobian of the closed-loop dynamics $f$ over the state box $B$ and parameter bin $\Lambda_k$.
If certification fails, a stack-based branch-and-bound recursively bisects the state box $B$ to narrow interval widths until either the condition holds or a maximum depth is reached.
Since bisection only adjusts interval widths rather than the candidate $P$, the resulting proof holds globally, yielding a certified metric $P_{\text{best}}$ and sublevel $c_{\text{best}}$ for the contraction ellipsoid $\mathcal{E}$ that guarantees stability for $C_{HA}(x)$ across the entire parameter envelope.

Comparing the contraction ellipsoid $\mathcal{E}$ obtained for the exemplary path tracking problem (\cref{fig:roa_intersection}) to the overall \ac{roa} highlights the conservatism that motivates our second-stage reachability analysis.

\subsubsection{Stage 2 -- Per-bin reachability bisection}
\label{subsubsec:roa_stage_2}

\begin{algorithm}[t]
\caption{Design Time \ac{roa} Identification}
\label{alg:roa_calculation}
\footnotesize
\begin{algorithmic}[1]
\REQUIRE \ac{ha} controller $C_{HA}$, closed-loop dynamics $f(x, C_{HA}(x), \lambda)$, state domain $\mathcal{X}$, parameter space partition $\Lambda_P$, safety set $\mathcal{X}_\mathcal{S}$, contraction metric $P$, sublevel value $c$ (defining contraction ellipsoid $\mathcal{E}$)
\ENSURE Recovery profile $\mathcal{R} = (P, c, \{\mathcal{A}_{\Lambda_k}\}_{k=1}^K, \tau_{d_{\text{max}}})$
\STATE $\tau_{d_{\text{max}}} \leftarrow 0$ \COMMENT{Maximum dwell time}
\FORALL{parameter bin $\Lambda_k \in \Lambda_P$} \label{algline:stage2_loop}
    \STATE $\mathcal{A}_{\Lambda_k} \leftarrow \emptyset$
    \STATE $W \leftarrow \{ \text{initial cell } \mathcal{X} \}$
    \WHILE{$W \neq \emptyset$}
        \STATE $\mathcal{Q} \leftarrow W.\text{pop}()$
        \STATE \colorbox{gray!20}{$\Phi \leftarrow \text{TMJets}(f, \mathcal{Q}, \Lambda_k, C_{HA}, T_{\text{reach}})$} \COMMENT{Reachability}
        \IF{$\Phi \subseteq \mathcal{X}_\mathcal{S}$ \AND $\exists t \le T_{\text{reach}} : \Phi(t) \subseteq \mathcal{E}$}  \label{algline:verify_cell}
            \STATE $t_{\text{entry}} \leftarrow \min \{ t \mid \Phi(t) \subseteq \mathcal{E} \}$
            \STATE $\tau_{d_{\text{max}}} \leftarrow \max(\tau_{d_{\text{max}}}, t_{\text{entry}})$
            \STATE $\mathcal{A}_{\Lambda_k} \leftarrow \mathcal{A}_{\Lambda_k} \cup \{\mathcal{Q}\}$
        \ELSIF{$\text{depth}(\mathcal{Q}) < d_{\text{max}}$}
            \STATE $W.\text{push}(\text{Split}(\mathcal{Q}))$
        \ENDIF
    \ENDWHILE
\ENDFOR
\RETURN Recovery profile $\mathcal{R} = (P, c, \{\mathcal{A}_{\Lambda_k}\}_{k=1}^K, \tau_{d_{\text{max}}})$
\end{algorithmic}

\end{algorithm}

The second stage reduces this conservatism through reachability analysis, \cmpr{subsec:soa_rta,subsubsec:reachability}.
Since reachability is equally affected by the wrapping effect of set propagation, we partition $\Lambda$ into bins $\Lambda_k \in \Lambda_P$ as before, turning the system into a switched system whose modes are the individual partitions.
For each mode, \cref{alg:roa_calculation} computes a recovery profile $\mathcal{R}$ comprising the set of safe \acp{roa} $\{\mathcal{A}_{\Lambda_k}\}_{k=1}^K$, the overall maximum dwell time $\tau_{d_{\text{max}}}$, and the Stage~1 ellipsoid $\mathcal{E}$ (represented by $P$ and $c$).

Starting from the entire state space $\mathcal{X}$, the algorithm applies Taylor-model forward reachability~\mycite{althoff2021set,dongxu2020reachability} over a horizon $T_{\text{reach}}$ and verifies whether the resulting reach-tube $\Phi$: (i)~remains entirely within the safe set, i.e., $\Phi(t) \subseteq \mathcal{X}_\mathcal{S}$ for all $t \in [0, T_{\text{reach}}]$ (safety verification), and (ii)~eventually enters the global contraction ellipsoid, i.e., $\exists t \le T_{\text{reach}} : \Phi(t) \subseteq \mathcal{E}$ (stability verification).
If this check returns successfully, we record the time needed to enter the contraction ellipsoid as the dwell time.
Otherwise, we bisect the state space into smaller cells $\mathcal{Q}$ and recurse until a maximum depth of $d_{\text{max}}$ is reached again.
This accumulates cells $\mathcal{Q}$ for which stability and safety are guaranteed through per-trajectory proofs.

\Cref{fig:ha_sim} shows a reach-tube for the path-tracking use case, entering $\mathcal{E}$ while maintaining the safety requirement $|e_y| \leq \SI{1}{\meter}$.
The Taylor-model bounds initially diverge, due to the interval representation, but ultimately contract into $\mathcal{E}$.
\cref{fig:roa_intersection} shows the pre-computed safe \ac{roa} $\mathcal{A}_{\Lambda_k}$ for the curvature bin $\kappa_{\text{ref}} \in [0.85, 0.95]$ along with the region associated with $\kappa_{\text{ref}} \in [-0.95, -0.85]$.
The two \acp{roa} overlap substantially but differ in coverage, since the associated curvatures determine whether $C_{HA}$ can safely stabilize the system.
Wider $\kappa_{\text{ref}}$ bins would make the approximation more conservative and reduce the distinction in coverage differences.
Conversely, to exploit coverage variation, we aim to construct a context-dependent \ac{roa} at runtime.

\subsubsection{Runtime Construction of Context-Dependent \acp{roa}}
\label{subsubsec:roa_runtime}

While global \acp{roa} provides unconditional stability over the entire parameter space $\Lambda$ (\cmpr{subsec:soa_roa}), forward reachability restricts the proof to the currently relevant context (\cmpr{subsec:soa_rta}).
The recovery profile $\mathcal{R}$ enables leveraging both.
To that end, we assume access to contextual knowledge of the active parameter subset $\Lambda_{\text{runtime}} \subseteq \Lambda$ valid over the maximum dwell time $\tau_{d_{\text{max}}}$. 
Given the bin partition $\{\Lambda_k\}_{k=1}^K$, the active parameters lie within the union of touched bins $\Lambda_{\text{runtime}} \subseteq \bigcup_{k \in \mathcal{I}} \Lambda_k$, indexed by the set $\mathcal{I}$ symbolizing runtime contextual knowledge.
To guarantee safety under all possible parameter variations in $\Lambda_{\text{runtime}}$, the state must belong to the safe \ac{roa} of all active bins.
Set theory then dictates that the context-dependent safe \ac{roa} $\mathcal{A}_{\text{runtime}}$ is constructed through intersection:
\begin{equation}
\label{eq:roa_runtime}
\begin{split}
	\mathcal{A}_{\text{runtime}} &= \bigcap_{k \in \mathcal{I}} \mathcal{A}_{\Lambda_k}, \\
	\text{where} \quad \mathcal{I} &= \{ k \in \{1,\dots,K\} \mid \Lambda_k \cap \Lambda_{\text{runtime}} \neq \emptyset \}
\end{split}
\end{equation}
By defining $\mathcal{A}_{\text{runtime}}$ as this intersection, safe recovery to $\mathcal{E}$ is guaranteed under parameter transitions within $\tau_{d_{\text{max}}}$ as each cell constitutes a trajectory-based proof.

Furthermore, the reduction in conservatism is demonstrated mathematically by the fact that restricting the intersection to active bins yields a larger \ac{roa}, since $\mathcal{I} \subseteq \{1, \dots, K\}$ implies that the context-dependent $\mathcal{A}_{\text{runtime}}$ is a superset of the global intersection ($\bigcap_{k=1}^K \mathcal{A}_{\Lambda_k} \subseteq \mathcal{A}_{\text{runtime}}$).
As shown in \cref{fig:roa_intersection}, a global \ac{roa} is limited to the small dark green intersection, whereas our approach dynamically leverages the larger context-dependent intersection.
In contrast, since calculating the conditional safe \ac{roa} requires only the intersection of sets, computational overhead is limited, making the approach suitable for runtime application, but at the cost of being bound to the parameter space $\Lambda$ defined at design time.

\subsection{Decision Module}
\label{subsec:decision_module}

Based on the recovery profile $\mathcal{R}$ and the runtime construction of context-dependent \acp{roa}, we now derive the switching rules of the decision module.
Rather than assuming a single \ac{ha} controller, however, we aim to further reduce conservatism by allowing multiple \ac{ha} candidates that may uniquely cover distinct regions of the state and parameter spaces.
This idea influences the design of the decision module in two ways.
First, the \ac{hp}-to-\ac{ha} switch (\cref{subsubsec:hp_to_ha_switch}) leverages that \textit{any} candidate certifying a proposed \ac{hp} action is sufficient to guarantee safety.
Second, we introduce the \ac{ha}-to-\ac{ha} switch (\cref{subsubsec:ha_to_ha_switch}), which enables dynamic switching between candidates to optimize recovery performance as the system state changes.
Finally, \cref{subsubsec:ha_to_hp_switch} describes how control is safely handed back to the \ac{hp} controller once recovery is complete.

\subsubsection{\ac{hp} to \ac{ha} Switch}
\label{subsubsec:hp_to_ha_switch}

\begin{algorithm}[t]
\caption{HP to HA Switching Logic}
\label{alg:hp_to_ha_switch}
\footnotesize
\begin{algorithmic}[1]
\REQUIRE State $x$, \ac{hp} action $u$, parameter space $\Lambda$, \ac{ha} candidates $\{C_{HA,j}\}_{j=1}^{M}$ with profiles $\{\mathcal{R}_j\}_{j=1}^{M}$ sorted ascending by $\tau_{d_{\text{max}}}$
\ENSURE Certifying candidate index $i^* \in \{1,\dots,M\} \cup \{\emptyset\}$

\STATE $x' \leftarrow \text{PredictState}(x, u, T_{\text{pred}})$
\FORALL{candidate $j \in \{1,\dots,M\}$}
    \IF{$\text{IsAdmissible}(x', \Lambda, \mathcal{R}_j)$}
        \RETURN $j$ \COMMENT{Safe candidate found (\cmpr{alg:is_admissible})}
    \ENDIF
\ENDFOR

\RETURN $\emptyset$ \COMMENT{No candidate certifies safety, switch to HA}
\end{algorithmic}

\end{algorithm}
To formalize the use of multiple \ac{ha} candidates, we generalize the \ac{ha} controller $C_{HA}(x)$ to a set $\{C_{HA,i}(x)\}_{i=1}^{M}$, each equipped with its own recovery profile $\mathcal{R}_i$ from \cref{alg:roa_calculation}.
Then, \cref{alg:hp_to_ha_switch} decides whether to switch to \ac{ha} mode by sequentially evaluating all candidates using the admissibility check of \cref{alg:is_admissible}.
Note that we sort the candidates by $\tau_{d_{\text{max}}}$ to select the one offering the fastest recovery, thereby maximizing the execution of the nominal \ac{hp} controller. 
However, other performance criteria could be optimized alternatively.

\begin{algorithm}[t]
\caption{Admissibility Check for HA Candidates}
\label{alg:is_admissible}
\footnotesize
\begin{algorithmic}[1]
\REQUIRE Predicted state $x'$, parameter space $\Lambda$, candidate recovery profile $\mathcal{R}_j = (P_j, c_j, \{\mathcal{A}_{j,k}\}_{k=1}^K, \tau_{d_{\text{max}},j})$
\ENSURE Result $\in \{\TRUE, \FALSE\}$

\STATE $\Lambda_{\text{runtime}} \leftarrow \textsc{ActiveParameterSpace}(\Lambda, \tau_{d_{\text{max}},j})$ 
\STATE $\mathcal{A}_{\text{runtime}} \leftarrow \textsc{IntersectRoAs}(\{\mathcal{A}_{j,k}\},\Lambda_{\text{runtime}}) $ \COMMENT{\cref{eq:roa_runtime}}
\IF{$x' \in \mathcal{A}_{\text{runtime}}$}
    \RETURN \TRUE
\ENDIF
\RETURN \FALSE
\end{algorithmic}

\end{algorithm}
For checking admissibility, \cref{alg:is_admissible} estimates the active parameter subset $\Lambda_{\text{runtime}}$ valid over the candidate's recovery time horizon $\tau_{d_{\text{max}},j}$ and applies \cref{eq:roa_runtime} to obtain the context-dependent runtime safe \ac{roa} $\mathcal{A}_{\text{runtime}}$.
If $x' \in \mathcal{A}_{\text{runtime}}$, the candidate is admissible and certifies safety of the proposed action because we have a trajectory-based proof that the \ac{ha} candidate will be able to recover the system when switched to it in the predicted state.
This also guarantees that the system will safely recover when switching to the last returned index after receiving the empty set from \cref{alg:hp_to_ha_switch}, which triggers the \ac{hp} to \ac{ha} switch.

In the path-tracking use case, we will leverage multiple \ac{rpp} controllers with varying configurations (e.g., velocity, lookahead distance) as different \ac{ha} controllers, \cmpr{subsec:ha_search}.

\subsubsection{\ac{ha} to \ac{ha} Switch}
\label{subsubsec:ha_to_ha_switch}

\begin{algorithm}[t]
\caption{HA to HA Switching Logic}
\label{alg:ha_to_ha_switch}
\footnotesize
\begin{algorithmic}[1]
\REQUIRE State $x$, active candidate index $i$, candidates $\{C_{HA,j}\}_{j=1}^{M}$ with profiles $\{\mathcal{R}_j\}_{j=1}^{M}$ sorted ascending by $\tau_{d_{\text{max}}}$, active action $u = C_{HA,i}(x)$, runtime parameter space $\Lambda_{\text{runtime}}$
\ENSURE Updated active candidate index $i^*$

\STATE $x' \leftarrow \text{PredictState}(x, u, T_{\text{pred}})$
\FORALL{candidate $j$ with $\tau_{d_{\text{max}},j} < \tau_{d_{\text{max}},i}$}
    \IF{$\text{IsAdmissible}(x', \Lambda_{\text{runtime}}, \mathcal{R}_j)$}
        \RETURN $j$ \COMMENT{Transition to faster recovery (\cmpr{alg:is_admissible})}
    \ENDIF
\ENDFOR
\RETURN $i$ \COMMENT{Maintain active candidate}
\end{algorithmic}

\end{algorithm}
While the \ac{hp} to \ac{ha} switch selects the last certifying candidate that yields the fastest recovery, others may become admissible as the state converges.
To optimize recovery, the decision module enables an \ac{ha}-to-\ac{ha} switch (\cref{alg:ha_to_ha_switch}).
For that, however, the next state $x'$ is predicted under the active candidate $C_{HA,i^*}(x)$ instead of using the proposed \ac{hp} action.
Transitioning to another candidate $j$ is then permitted only if $x'$ is certified admissible, which requires re-evaluating the active parameter space over the lookahead horizon.
This ensures the new controller takes over from a state within its certified \ac{roa}.
Since candidates are finite and sorted, the number of switches per recovery phase is finite, preventing chattering.

In the path-tracking use case, we restrict transitions to when the robot is heading toward the path $\mathcal{P}$ (i.e., $\operatorname{sgn}(e_y) \neq \operatorname{sgn}(e_\theta)$), which prioritizes restricting the maximal path deviation next to the recovery time.

\subsubsection{\ac{ha} to \ac{hp} Switch}
\label{subsubsec:ha_to_hp_switch}

\begin{algorithm}[t]
\caption{HA to HP Switching Logic}
\label{alg:ha_to_hp_switch}
\footnotesize
\begin{algorithmic}[1]
\REQUIRE State $x$, active index $i^*$, candidate profiles $\{\mathcal{R}_j\}_{j=1}^{M}$, hysteresis ratio $\alpha$
\ENSURE Result $\in \{\TRUE, \FALSE\}$

\STATE $V(x) \leftarrow x^T \mathcal{R}_{i^*}.P x$ \COMMENT{Evaluate Lyapunov function around equilibrium}
\IF{$V(x) \le \alpha \cdot \mathcal{R}_{i^*}.c$}
    \RETURN \TRUE \COMMENT{Safe to return control to HP}
\ENDIF
\RETURN \FALSE \COMMENT{Remain in HA mode}
\end{algorithmic}

\end{algorithm}
While in \ac{ha} mode, the goal is to recover the system to a state where control can be safely handed back to the \ac{hp} controller $C_{HP}(x)$.
Instead of imposing conservative static dwell times, we monitor the system state and switch back once it enters the contraction ellipsoid $\mathcal{E}_{i^*}$ (\cref{alg:ha_to_hp_switch}).
This leverages the contraction metric $P_{i^*}$, which acts as a common Lyapunov function at the nominal equilibrium.
With that, switched systems theory guarantees stability under arbitrary switching once we entered the ellipsoid, making dwell times unnecessary.

To further prevent local boundary chattering, we introduce a hysteresis ratio $\alpha \in (0,1)$ and trigger the switch only when the Lyapunov function $V(x) = x^T P_{i^*} x \le \alpha \cdot c_{i^*}$.

For the path-tracking use case, the safe \ac{roa} is symmetric around the nominal equilibrium $[e_y, e_\theta]^T = [0, 0]^T$ (\cref{fig:roa_intersection}), making hysteresis a formal safety margin. 
However, generic use cases may require $\alpha$ to prevent chattering in case of non-symmetric \acp{roa}.

\subsubsection{Complete Simplex Controller}
\label{subsubsec:complete_simplex_controller}

\begin{algorithm}[t]
\caption{Complete Simplex Controller Logic}
\label{alg:simplex_logic}
\footnotesize
\begin{algorithmic}[1]
\REQUIRE State $x$, parameter space $\Lambda$, candidates $\{C_{HA,j}\}_{j=1}^{M}$
\ENSURE Control action $u$

\IF{\NOT in\_ha\_mode}
    \STATE $u \leftarrow C_{HP}(x)$
    \STATE $i^* \leftarrow \textsc{HPtoHA}(x, u, \Lambda, \{C_{HA,j}\}_{j=1}^{M})$ %
    \IF{$i^* \neq \emptyset$}
        \STATE active\_index $\leftarrow i^*$
        \RETURN $u$
    \ELSE
        \STATE in\_ha\_mode $\leftarrow \TRUE$
        \RETURN $C_{HA, \text{active\_index}}(x)$
    \ENDIF
\ELSE
    \STATE $u \leftarrow C_{HA, \text{active\_index}}(x)$
    \IF{$\textsc{HAtoHP}(x, \text{active\_index})$}
        \STATE in\_ha\_mode $\leftarrow \FALSE$ %
        \RETURN $C_{HP}(x)$
    \ENDIF

    \STATE $u \leftarrow C_{HA, \text{active\_index}}(x)$
    \STATE active\_index $\leftarrow \textsc{HAtoHA}(x, \text{active\_index}, \Lambda_{\text{runtime}})$ %
    \RETURN $u$
\ENDIF
\end{algorithmic}
\end{algorithm}

The combination of the three switching rules into the final Simplex controller $C_S(x)$ is summarized in \cref{alg:simplex_logic}. 
It illustrates the execution during a single control cycle. 
If the system is in the nominal \ac{hp} mode, the \ac{hp} controller $C_{HP}(x)$ calculates a proposed control action $u$, which is verified against all \ac{ha} candidates using the \ac{hp}-to-\ac{ha} switching logic (\cref{alg:hp_to_ha_switch}). 
If the action is deemed safe by any candidate $i^*$, it becomes the active candidate $C_{HA,i^*}(x)$ for a potential future recovery, and the \ac{hp} action is applied. 
If no candidate certifies safety, the mode switches to \ac{ha}, and the active candidate $C_{HA,i^*}(x)$ is invoked to recover the system. 
During the \ac{ha} mode, the system monitors if recovery into the contraction ellipsoid is achieved via the \ac{ha}-to-\ac{hp} logic (\cref{alg:ha_to_hp_switch}), returning control to the \ac{hp} controller $C_{HP}(x)$. 
If not fully recovered, the \ac{ha}-to-\ac{ha} logic (\cref{alg:ha_to_ha_switch}) evaluates whether transitioning to another \ac{ha} candidate $C_{HA,j}(x)$ optimizes the ongoing recovery. 
The active \ac{ha} candidate then issues the next control action.

\section{Evaluation}
\label{sec:evaluation}

The previous section detailed our multi-candidate Simplex controller, which we now evaluate against three claims: that the decision module maintains the lateral safety requirement under fault injection, that switching between multiple \ac{ha} candidates enlarges the permissible operating region of the uncertified \ac{hp} controller, and that its computational overhead is compatible with real-time execution on embedded hardware.

For that, we proceed in three stages.
First, \ac{ode}-based simulation (\cref{subsec:ode_eval}) enables using the same dynamics as used for the \ac{roa} approximation and thereby allows quantifying the benefit of multi-candidate switching and, most importantly, showing that a one-step prediction is, in fact, guaranteeing safety.
Gazebo simulation (\cref{subsec:gazebo_eval}) introduces unmodeled dynamics and allows identification of the prediction horizon required to maintain safety. 
Real-world experiments (\cref{subsec:real_eval}) confirm that this identified prediction horizon is close to reality, quantify the decision module's overhead, and demonstrate the shielding of a behavioral-cloning \ac{hp} controller.

As a prerequisite, \cref{subsec:prototype} describes the mobile robot platform, and \cref{subsec:ha_search} details how \cref{alg:roa_calculation} was used to search for suitable \ac{ha} controllers, motivating the two \ac{rpp} configurations used throughout the evaluation.

\subsection{Description of the Mobile Robot}
\label{subsec:prototype}

We use a custom differential-drive mobile robot designed for urban navigation tasks, equipped with an Intel Core i7-12700TE CPU (35\,W TDP). 
The platform's hardware limits the control actions to a linear velocity $v \in [0, v_{\text{max}}]$ with $v_{\text{max}} = \SI{1.5}{\meter\per\sec}$ and an angular velocity $\omega \in [-\omega_{\text{max}}, \omega_{\text{max}}]$ with $\omega_{\text{max}} = \SI{6.5}{\radian\per\sec}$.
Motivated by a real-world parcel delivery scenario %
where the robot must navigate a sidewalk, we define a strict lateral safety margin of $\SI{1}{\meter}$. 
This establishes the safety requirement $\mathcal{X}_\mathcal{S} = \left\{|e_y| \leq \SI{1}{\meter}\right\}$ during path tracking.
Although this margin is scenario-specific, different requirements can be adopted without affecting the underlying methodology.

\subsection{Searching for \ac{ha} controllers}
\label{subsec:ha_search}

Before deploying our Simplex architecture, we must identify and certify a set of safe \ac{ha} controllers.
For that, we first detail the employed \ac{rpp} algorithm (\cmpr{subsubsec:ha_controller_design}) before \cref{subsubsec:curvature_estimation_tradeoffs} reports on the runtime context identification, that is, curvature approximation.
Finally, we present the application of our two-stage pipeline to identify the recovery profile for two specific \ac{rpp} configurations we apply in this evaluation.

\subsubsection{High-Assurance Controller Design}
\label{subsubsec:ha_controller_design}

Given that calculating the \ac{ha}'s recovery profile involves interval propagation prone to wrapping effects, we favor a simple control law to mitigate this effect.
Thus, we employ the geometric \ac{rpp} controller $C_{RPP,\gamma}(x)$ with configuration \mbox{$\gamma = [L_{\min}, t_h, R_{\min}, v_{\text{des}}]^T$}, derived from \mycite{macenski2023regulated}.
At each control step, \ac{rpp} regulates the lookahead distance $L$ and calculates control actions $v, \omega$:
\begin{align}
	L &= \sqrt{L_{\min}^2 + (t_h \cdot v_{\text{steady}})^2}, \quad v_{\text{steady}} = \frac{v_{\text{des}}}{\sqrt{1 + (R_{\min} \kappa_{\text{ref}})^2}} \label{eq:rpp_lookahead} \\
	v &= \frac{v_{\text{des}}}{\sqrt{1 + (R_{\min} \kappa_{\text{target}})^2}}, \quad \omega = v \kappa_{\text{target}} \label{eq:ha_control}
\end{align}
where $\kappa_{\text{target}}$ is the curvature of the target arc to the lookahead point. 
Crucially, unlike standard \ac{rpp}~\mycite{macenski2023regulated} which regulates $L$ using the current robot velocity $v_{\text{robot}}$, we regulate $L$ using the curvature-dependent steady-state velocity $v_{\text{steady}}$. 
This removes the circular dependency between $L$ and $v$, which would otherwise amplify the wrapping effect during interval propagation.

\subsubsection{Curvature Estimation}
\label{subsubsec:curvature_estimation_tradeoffs}

To construct the context-dependent \ac{roa} at runtime, the decision module estimates the reference curvature parameter space $\Lambda_{\text{runtime}} = [\kappa_{\text{lo}}, \kappa_{\text{hi}}]$.
For that, we fit a circle $(x-a)^2 + (y-b)^2 = R^2$ to a sliding window of waypoints $j \in [i - N_{\text{circ}}, i + N_{\text{circ}}]$ (clamped to path boundaries) using linear least-squares.
This yields the curvature magnitude $\kappa_{\text{mag}} = 1/R$, with its sign determined by the cross product of successive chord vectors.

Note that the choice of $N_{\text{circ}}$ involves a safety-critical trade-off: a large window ($N_{\text{circ}}=30$) filters sensor and discretization noise but can under-approximate local curvatures at sharp transitions, leading to false negatives.
Conversely, a small window ($N_{\text{circ}}=5$) tracks transitions accurately but is highly sensitive to noise, triggering false-positive interventions.
Thus, we use $N_{\text{circ}}=5$ for the idealized ODE simulations (\cref{subsec:ode_eval}), but adopt $N_{\text{circ}}=30$ in Gazebo and physical experiments to prevent noise-induced false positives.

\subsubsection{Configuration Search}
\label{subsubsec:offline_verification_sweep}

For applying the Simplex architecture, we want to (i) maximize the \ac{roa} of the \ac{ha} controller and (ii) identify multiple, complementary \ac{ha} candidates.
Thus, we used the process described in \cref{subsec:roa} to identify the recovery profile for multiple configurations and chose the one that yielded the highest coverage and the highest desired velocity.
Specifically, we constructed 24 configurations: $v_{\text{des}} \in \{\SI{1}{\meter\per\sec}, \SI{1.5}{\meter\per\sec}\}$, $L_{\text{min}} \in \{\SI{0.35}{\meter}, \SI{0.7}{\meter}\}$, $t_h \in \{\SI{0.35}{\sec}, \SI{0.7}{\sec}, \SI{1.0}{\sec}\}$, and $R_{\min} \in \{\SI{0.3}{\meter}, \SI{0.7}{\meter}\}$.
For each candidate, we sampled $N_P = 10$ candidates and bisected cells up to depth $d_{\text{max}} = 14$ to establish the contraction ellipsoid $\mathcal{E}$, \cmpr{alg:contraction_analysis}.
We then applied forward reachability (\cref{alg:roa_calculation}) over $T_{\text{reach}} = \SI{3.5}{\sec}$ with bisection depth limit $d_{\text{max}} = 10$, using the TMJets algorithm configured with time order $orderT = 8$ and jet transport order $orderS = 2$ to obtain the final recovery profile~\mycite{JuliaReach19}.

\begin{table}[ht]
\centering
\caption{Summary of \ac{roa} Calculations}
\label{tab:reachability_summary_full}
{\footnotesize
\begin{tabular}{lrrrrr}
\toprule
$\gamma_i$ & $v_\text{des}$ [m/s] & $L_\text{min}$ [m] & $t_h$ [s] & $R_\text{min}$ [m] & Safe Cells \\
\midrule
$\gamma_{4}$ & 1.00 & 0.35 & 1.00 & 0.30 & 752  \\
$\gamma_{15}$ & 1.50 & 0.70 & 0.70 & 0.30 & 664 \\
\bottomrule
\end{tabular}
}

\end{table}
To compare configurations, we intersected the individual \acp{roa} across the entire $\kappa_{\text{ref}}$ range and counted the number of cells deemed safe.
While $\gamma_4$ achieved the highest number of safe cells with $v_{\text{des}} = \SI{1.0}{\meter\per\sec}$, $\gamma_{15}$ reached the highest number with $v_{\text{des}} = \SI{1.50}{\meter\per\sec}$, \cmpr{tab:reachability_summary_full}.
The offline verification of a single configuration took approximately \SI{100}{\minute}~\SI{37}{\second} of wall-clock time on a laptop with an Intel Core Ultra~7~155H CPU (22 threads) and \SI{64}{\giga\byte} of RAM.
\cref{fig:roa_intersection} shows two \ac{roa} obtained for $\gamma_4$.

\subsection{ODE-based Simulation}
\label{subsec:ode_eval}

This first stage evaluates whether a one-step prediction $T_{\text{pred}} = \tau_C$ in \cref{alg:hp_to_ha_switch} suffices to guarantee safety under idealized dynamics.
We simulate the error dynamics \cref{eq:error_dynamics} via Runge-Kutta-Fehlberg 4/5 along random reference paths for which we concatenate segments with uniformly distributed lengths $s_{\text{seg}} \in \mathbb{U}(\SI{1.0}{\meter}, \SI{4.0}{\meter})$ and curvatures $\kappa_{\text{seg}} \in \mathbb{U}(\SI{-0.95}{\per\meter}, \SI{0.95}{\per\meter})$.
To align with our prototype, we implement a discrete-time control period of $\tau_C = \SI{0.05}{\sec}$ where control actions $u = C_S(x)$ are held constant, relaxing the continuous control assumption from \cref{sec:concept}.

To validate our approach, we perform four Monte-Carlo campaigns ($N = 1000$ runs each) with the Simplex controller $C_S(x)$, using the configured \ac{ha} controller also as the uncertified \ac{hp} controller. 
We evaluate: (1) the single \ac{ha} candidate $C_{RPP,\gamma_4}$, (2) the single candidate $C_{RPP,\gamma_{15}}$, (3) multi-candidate switching between $C_{RPP,\gamma_4}$ and $C_{RPP,\gamma_{15}}$, and (4) campaign (3) with the path alignment check during \ac{ha} to \ac{ha} switching disabled. 
The desired velocity $v_{\text{des}}$ is $\SI{1.0}{\meter\per\sec}$ for (1) and $\SI{1.5}{\meter\per\sec}$ for (2)--(4).
In all runs, we injected a fault for $t \in [\SI{2.0}{\sec}, \SI{5.0}{\sec}]$ by inverting the \ac{hp} controller's angular velocity ($\omega_{HP} = -\omega_{RPP}$), forcing the decision module to detect the violation and switch to the active, fault-free \ac{ha} candidate.

\begin{figure}
    \input{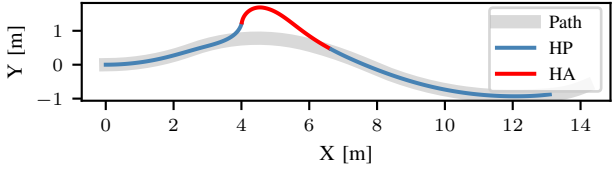}
    \caption{Exemplary trajectory during \ac{ode}-based simulation for $C_{RPP, \gamma4, \gamma_{15}}$ showing recovery into the safe \ac{roa}}
    \label{fig:traj_ivp}
\end{figure}
In the exemplary run (\cref{fig:traj_ivp}), the injected fault drives the vehicle under \ac{hp} control (blue) off the reference path (gray) until the decision module switches to the \ac{ha} controller (red) to restore safety. During this intervention, the path turns away from the robot, representing a worst-case recovery.

\begin{table}[t]
\centering
\caption{Summary of ODE-based Monte-Carlo Simplex Simulation Results ($N = 1000$ Runs per Row)}
\label{tab:ode_sim_summary}
{\footnotesize
\setlength{\tabcolsep}{1.0pt}
\begin{tabular}{@{} lccccccc @{}}
\toprule
\multirow{2}{*}{\textbf{Config.}} & \multicolumn{3}{c}{$T_{\text{HA}}$ [s]} & \multicolumn{3}{c}{$|e_y|_{\text{switch}}$ [m]} & \multirow{2}{*}{$|e_y|_{\text{max}}$ [m]} \\
\cmidrule(r){2-4} \cmidrule(l){5-7}
 & \textbf{Min} & \textbf{Mean} & \textbf{Max} & \textbf{Min} & \textbf{Mean} & \textbf{Max} &  \\
(1) $\gamma_4$ & 0.500 & 2.819 & 4.450 & 0.001 & 0.406 & 0.601 & 0.982 \\
(2) $\gamma_{15}$ & 0.700 & 2.337 & 4.450 & 0.000 & 0.392 & 0.656 & 0.964 \\
(3) $\{\gamma_4, \gamma_{15}\}$ & 0.800 & 2.463 & 4.400 & 0.001 & 0.439 & 0.706 & 0.967 \\
\midrule
(RO) $\{\gamma_4, \gamma_{15}\}$ & 0.450 & 2.414 & 4.300 & 0.001 & 0.438 & 0.678 & 0.977 \\
\bottomrule
\end{tabular}
}

\end{table}
The aggregate statistics in \cref{tab:ode_sim_summary} compile the recovery duration ($T_{\text{HA}}$), lateral deviation at switching ($|e_y|_{\text{switch}}$), and maximum path deviation ($|e_y|_{\text{max}}$).
Since all $|e_y|_{\text{max}} \leq \SI{1.0}{\meter}$, the controller guarantees safety, confirming that a one-step prediction is sufficient when model dynamics are matched.

Comparing the setups regarding $T_{\text{HA}}$, $\gamma_4$ has the longest mean recovery but transitions conservatively at a large mean deviation of $\SI{0.406}{\meter}$.
Conversely, $\gamma_{15}$ recovers fastest on average but has a smaller \ac{roa}, which forces earlier switching and thus restricts the \ac{hp} controller's operating region.
Multi-candidate switching combines these strengths: starting recovery with $\gamma_4$ and transitioning to $\gamma_{15}$ once the error decreases maximizes the allowable nominal boundary ($|e_y|_{\text{switch}} = \SI{0.439}{\meter}$ mean, $\SI{0.706}{\meter}$ max) while achieving a fast recovery.
Lastly, disabling the heading gate alignment check (Experiment RO) yields a slightly faster mean recovery but increases the peak deviation to $|e_y|_{\text{max}} = \SI{0.977}{\meter}$ (compared to Experiment 3), confirming that the heading gate improves safety margins during \ac{ha}-to-\ac{ha} transitions without being strictly necessary from a theoretical perspective.

\subsection{Gazebo}
\label{subsec:gazebo_eval}

\begin{figure*}
    \centering
    \begin{subfigure}[b]{0.33\linewidth}
        \centering
        \includegraphics[width=\linewidth]{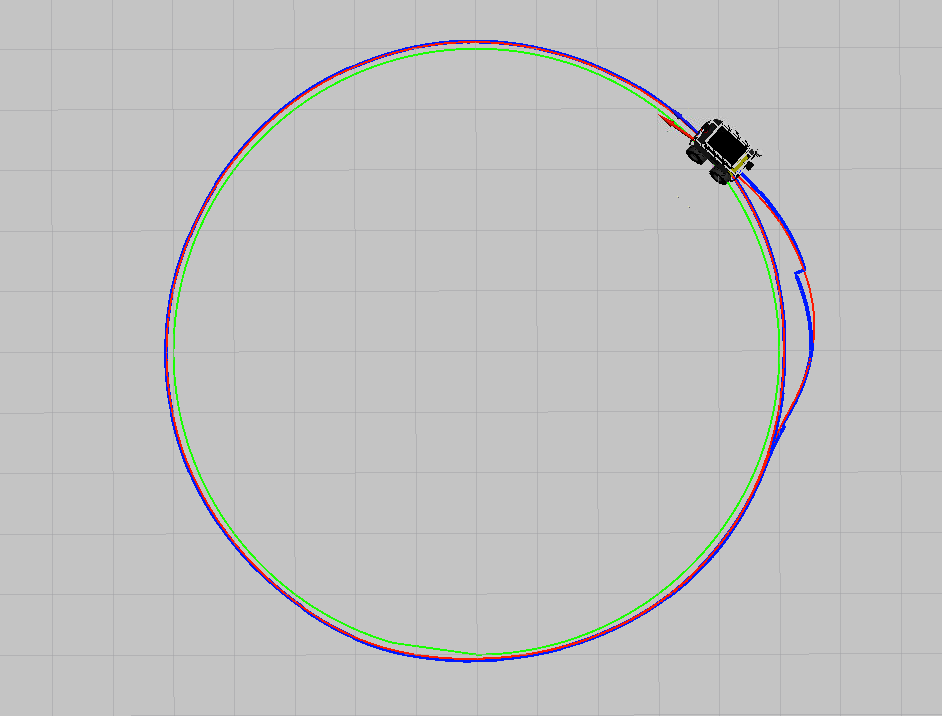}
        \caption{Circular path with $\kappa = 0.2$.}
        \label{fig:gazebo_circle}
    \end{subfigure}
    \hfill
    \begin{subfigure}[b]{0.6\linewidth}
        \centering
        \includegraphics[width=\linewidth]{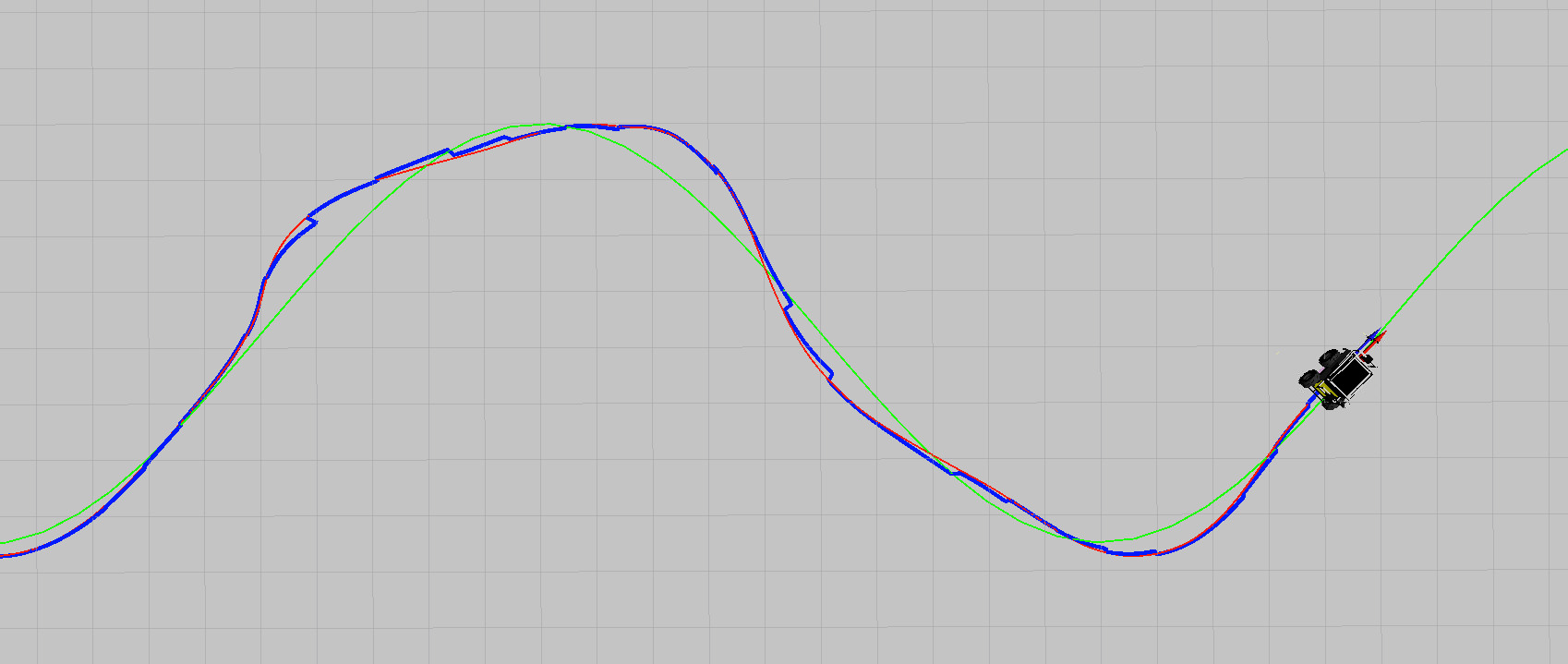}
        \caption{Synthetic cosine reference path.}
        \label{fig:gazebo_cosine}
    \end{subfigure}
    \caption{Gazebo simulation visualized by RViz2 with active failure injection. Green indicates the reference path, red is the ground truth provided by Gazebo, blue is the localization as seen by the robot.}
    \label{fig:gazebo_environments}
\end{figure*}

While the \ac{ode}-based evaluation confirmed safety under idealized dynamics, physical deployment introduces unmodeled dynamics and disturbances.
To evaluate robustness under these realistic conditions---specifically assessing the impact of both the prediction horizon $T_{\text{pred}}$ and the hysteresis threshold $\alpha$---we implement our Simplex controller in ROS2 Jazzy~\footnote{\url{https://docs.ros.org/en/jazzy/}} and simulate it in Gazebo.
The simulation employs a model with wheel slip (lateral and longitudinal slip compliance $= 1.0$, normal force \SI{113}{\newton} per wheel), Gaussian IMU angular velocity noise ($\sigma = \SI{e-3}{\radian\per\second}$ with bias), and undisturbed GNSS position updates at \SI{1}{\Hz}.
The robot first follows a circular reference path with $\kappa_{\text{ref}} = 0.2$ (\cref{fig:gazebo_circle}) before being deployed on a synthetic cosine-shaped reference path.
Using the configurations from \cref{subsec:ode_eval} under active fault injection, we run $N=10$ simulations per setup, varying the prediction horizon $T_{\text{pred}} = h \cdot \tau_C$ for $h \in \{1,2,5,10,15\}$ and the hysteresis factor $\alpha \in \{0.8, 0.9, 1.0\}$.
\Cref{tab:gazebo_sim_summary} reports the maximal absolute deviation $\max(|e_y|)$ from the path as well as the mean absolute deviation $\mu(|e_y|)$ during active \ac{ha} mode.

\begin{table}
\centering
\caption{Summary of Gazebo simulation for circular path ($\kappa_{\text{ref}} = 0.2$) with varying prediction horizons $h$ and hysteresis thresholds $\alpha$.}
\label{tab:gazebo_sim_summary}
{\footnotesize
\setlength{\tabcolsep}{2.0pt}
\begin{tabular}{@{}ll cc cc cc@{}}
\toprule
\multirow{2}{*}{\textbf{Config.}} & \multirow{2}{*}{$h$} & \multicolumn{2}{c}{$\alpha = 0.8$} & \multicolumn{2}{c}{$\alpha = 0.9$} & \multicolumn{2}{c}{$\alpha = 1.0$} \\
\cmidrule(lr){3-4} \cmidrule(lr){5-6} \cmidrule(lr){7-8}
 & & $\max(|e_y|)$ & $\mu(|e_y|)$ & $\max(|e_y|)$ & $\mu(|e_y|)$ & $\max(|e_y|)$ & $\mu(|e_y|)$ \\
\midrule
$\gamma_{15}$ & $1$ & 2.445 & 0.631 & 3.275 & 0.642 & 2.661 & 0.605 \\
$\gamma_{15}$ & $2$ & 1.690 & 0.544 & 1.657 & 0.566 & 1.693 & 0.553 \\
$\gamma_{15}$ & $5$ & 1.187 & 0.489 & 1.220 & 0.490 & 1.149 & 0.501 \\
$\gamma_{15}$ & $10$ & 0.774 & 0.314 & 0.670 & 0.295 & 0.812 & 0.367 \\
$\gamma_{15}$ & $15$ & 0.452 & 0.199 & 0.404 & 0.208 & 0.458 & 0.204 \\
\midrule
$\gamma_4$ & $1$ & 1.839 & 0.594 & 1.890 & 0.643 & 1.937 & 0.631 \\
$\gamma_4$ & $2$ & 1.467 & 0.521 & 1.429 & 0.538 & 1.642 & 0.569 \\
$\gamma_4$ & $5$ & 1.087 & 0.527 & 1.019 & 0.535 & 1.085 & 0.557 \\
$\gamma_4$ & $10$ & 0.703 & 0.426 & 0.681 & 0.424 & 0.689 & 0.424 \\
$\gamma_4$ & $15$ & 0.470 & 0.300 & 0.474 & 0.302 & 0.456 & 0.305 \\
\bottomrule
\end{tabular}
}

\end{table}
The simplified dynamics used to approximate the \acp{roa} do not capture the closer-to-real Gazebo dynamics, causing safety violations for both configurations $\gamma_{15}$ and $\gamma_4$ for prediction horizons up to $h=5$, whereas horizons of $h \ge 10$ successfully maintain the lateral safety requirement, \cmpr{tab:gazebo_sim_summary}.
A horizon of $h=15$ triggers conservative switches to the \ac{ha} controller, reducing peak deviations.
This data indicates that setting the prediction horizon to $T_{\text{pred}} = 10 \cdot \tau_C$ yields the best results. 

The hysteresis ratio $\alpha$, in contrast, has no systematic effect on tracking performance.
For $h=10$ (\cref{tab:gazebo_sim_summary}), the mean deviation $\mu(|e_y|)$ varies by at most \SI{0.07}{\meter} for $\gamma_{15}$ and \SI{0.02}{\meter} for $\gamma_4$ across all $\alpha$ values.
This is consistent with our hypothesis: the certified \acp{roa} are symmetric and cover the contraction ellipsoid $\mathcal{E}$ well, so the switching boundary is not tightly constrained by the contraction boundary, and the choice of $\alpha$ has negligible practical impact in this use case.

To test worst-case scenarios with $\gamma_4 + \gamma_{15}$ switching enabled, we simulate active fault injection on the outermost parts of a synthetic cosine path (\cref{fig:gazebo_cosine}).
With a prediction horizon of $h=10$, the maximum deviation across 10 runs remains bounded at $\SI{0.894}{\meter}$.
During 28 active \ac{ha} interventions, the average recovery duration under $\gamma_4$ was $\SI{0.906}{\second}$, compared to $\SI{5.934}{\second}$ under $\gamma_{15}$.
This disparity reflects the dynamic switching logic: starting with $\gamma_4$'s wider \ac{roa} and transitioning to the higher-velocity $\gamma_{15}$ once admissible.

\begin{figure*}[!t]
    \centering
    \begin{minipage}[t]{0.38\textwidth}
        \centering
        \resizebox{\columnwidth}{!}{\input{figures/generated/gazebo/cosine_faults/decision_module_9_h10_c4_c15_c4_c15_plot.pgf}}
        \caption{$C_{RPP, \gamma_4, \gamma_15}$ with $h=10$ on cosine path.}
        \label{fig:worst_case_c15_h10}
    \end{minipage}\hfill
    \begin{minipage}[t]{0.38\textwidth}
        \centering
        \resizebox{\columnwidth}{!}{\input{figures/generated/experiments/kappa_0_9/decision_module_7_plot.pgf}}
        \caption{$C_{RPP, \gamma_4 + \gamma_{15}}$ on circular track with $\kappa_{\text{ref}} = 0.9$.}
        \label{fig:circular_real_world_decision_module}
    \end{minipage}\hfill
    \begin{minipage}[t]{0.22\textwidth}
        \centering
        \resizebox{\columnwidth}{!}{\input{figures/generated/experiments/urz/h12/bc/track_2.pgf}}
        \caption{Real-world track.}
        \label{fig:real_world_traj}
    \end{minipage}
\end{figure*}

An exemplary simulation segment is shown in \cref{fig:worst_case_c15_h10}.
The non-smooth error signals $e_y$ and $e_\theta$ reflect simulated sensors, including $\SI{1}{\Hz}$ GNSS updates, as well as noise-affected IMU.
Centrally, the transition from \ac{ha}~0 ($\gamma_4$) to \ac{ha}~1 ($\gamma_{15}$) occurs at the peak lateral error, validating the heading admissibility condition.
Moreover, as \ac{ha} to \ac{ha} switches occur only when recovery time can be reduced, rapid chattering is prevented inherently despite noisy heading error values.

\subsection{Real-World Driving Scenarios}
\label{subsec:real_eval}

Having established the prediction horizon in Gazebo, we now validate whether $h=10$ maintains safety on the physical platform, quantify the magnitude of the remaining unmodeled dynamics, and provide a preliminary latency analysis.
We first repeat the circular scenarios for curvatures $\kappa_{\text{ref}} \in \{0.2, 0.5, 0.9\}$, then move to a realistic urban trajectory, testing it first under fault injection and then while shielding a behavioral-cloning \ac{hp} controller.

\subsubsection{Circular Scenarios}
\label{subsubsec:circular_scenarios}

We repeat the circular tracks with $v_{\text{des}} = \SI{1.0}{\meter\per\sec}$, randomly injecting $5$ to $10$ faults per configuration and curvature.
\Cref{tab:ha_real_deviations_merged} reports the results.

\begin{table}
    \centering
    \caption{Real-world path-tracking results under reference curvatures $\kappa_{\text{ref}} \in \{0.2, 0.5, 0.9\}$ across different prediction horizons $h$: $e_y^{\max}$ is the maximum lateral deviation (\si{\meter}), $N_{\text{a}}/N_{\text{v}}$ is the number of active \ac{ha} interventions / safety violations, and $\bar{T}_{\text{HA}}$ is the mean intervention duration (\si{\second}).}
    \label{tab:ha_real_deviations_merged}
    \resizebox{\columnwidth}{!}{{\footnotesize
\setlength{\tabcolsep}{0.8pt}
\begin{tabular}{@{}llccc@{\hspace{2pt}}ccc@{\hspace{2pt}}ccc@{}}
  \hline
  \textbf{Config.} & $h$ & \multicolumn{3}{c}{$\kappa_{\text{ref}} = 0.2$} & \multicolumn{3}{c}{$\kappa_{\text{ref}} = 0.5$} & \multicolumn{3}{c}{$\kappa_{\text{ref}} = 0.9$} \\
  & & $e_y^{\max}$ & $N_{\text{a}}/N_{\text{v}}$ & $\bar{T}_{\text{HA}}$ & $e_y^{\max}$ & $N_{\text{a}}/N_{\text{v}}$ & $\bar{T}_{\text{HA}}$ & $e_y^{\max}$ & $N_{\text{a}}/N_{\text{v}}$ & $\bar{T}_{\text{HA}}$ \\ \hline
  $\gamma_{15}$ & 1 & \cellcolor{red!10}5.40 & 6/2 & 5.3 & \cellcolor{red!10}7.20 & 11/4 & 4.9 & \cellcolor{red!10}9.46 & 3/3 & 19.6 \\
  $\gamma_{15}$ & 5 & \cellcolor{red!10}1.87 & 5/3 & 4.9 & \cellcolor{red!10}2.31 & 7/5 & 5.3 & \cellcolor{red!10}2.13 & 6/6 & 6.4 \\
  $\gamma_{15}$ & 10 & 0.94 & 28/0 & 3.1 & 0.87 & 11/0 & 2.1 & 0.62 & 6/0 & 8.8 \\
  $\gamma_{15}$ & 15 & 0.52 & 6/0 & 2.0 & 0.68 & 11/0 & 3.2 & 0.42 & 4/0 & 12.7 \\
  \hline
  $\gamma_4$ & 1 & \cellcolor{red!10}6.40 & 6/5 & 12.0 & \cellcolor{red!10}8.54 & 8/3 & 7.0 & \cellcolor{red!10}11.44 & 2/2 & 24.5 \\
  $\gamma_4$ & 5 & \cellcolor{red!10}1.42 & 5/3 & 3.9 & \cellcolor{red!10}1.32 & 14/6 & 2.1 & \cellcolor{red!10}1.69 & 5/4 & 4.8 \\
  $\gamma_4$ & 10 & 0.94 & 25/0 & 3.8 & 0.85 & 9/0 & 2.0 & 0.94 & 7/0 & 3.1 \\
  $\gamma_4$ & 15 & 0.57 & 7/0 & 2.2 & 0.46 & 10/0 & 2.9 & 0.55 & 8/0 & 2.8 \\
  \hline
   &  & \cellcolor{red!10} &  & $\gamma_4$: 8.4 & \cellcolor{red!10} &  & $\gamma_4$: 11.6 & \cellcolor{red!10} &  & $\gamma_4$: 14.9 \\
  \multirow{-2}{*}{$\gamma_4 + \gamma_{15}$} & \multirow{-2}{*}{1} & \cellcolor{red!10}\multirow{-2}{*}{9.16} & \multirow{-2}{*}{6/3} & $\gamma_{15}$: 1.0 & \cellcolor{red!10}\multirow{-2}{*}{8.76} & \multirow{-2}{*}{3/2} & $\gamma_{15}$: 4.8 & \cellcolor{red!10}\multirow{-2}{*}{10.82} & \multirow{-2}{*}{3/3} & $\gamma_{15}$: 5.8 \\
  \hline
   &  & \cellcolor{red!10} &  & $\gamma_4$: 1.4 & \cellcolor{red!10} &  & $\gamma_4$: 1.6 & \cellcolor{red!10} &  & --- \\
  \multirow{-2}{*}{$\gamma_4 + \gamma_{15}$} & \multirow{-2}{*}{5} & \cellcolor{red!10}\multirow{-2}{*}{1.46} & \multirow{-2}{*}{6/3} & $\gamma_{15}$: 4.6 & \cellcolor{red!10}\multirow{-2}{*}{1.96} & \multirow{-2}{*}{9/5} & $\gamma_{15}$: 4.5 & \cellcolor{red!10}\multirow{-2}{*}{1.72} & \multirow{-2}{*}{1/1} & $\gamma_{15}$: 62.6 \\
  \hline
   &  & \cellcolor{red!10} &  & $\gamma_4$: 1.2 &  &  & $\gamma_4$: 0.9 & \cellcolor{red!10} &  & $\gamma_4$: 0.9 \\
  \multirow{-2}{*}{$\gamma_4 + \gamma_{15}$} & \multirow{-2}{*}{10} & \cellcolor{red!10}\multirow{-2}{*}{1.04} & \multirow{-2}{*}{27/2} & $\gamma_{15}$: 3.2 & \multirow{-2}{*}{0.99} & \multirow{-2}{*}{8/0} & $\gamma_{15}$: 3.8 & \cellcolor{red!10}\multirow{-2}{*}{1.11} & \multirow{-2}{*}{8/1} & $\gamma_{15}$: 2.5 \\
  \hline
   &  &  &  & $\gamma_4$: 0.8 &  &  & $\gamma_4$: 1.0 &  &  & $\gamma_4$: 0.9 \\
  \multirow{-2}{*}{$\gamma_4 + \gamma_{15}$} & \multirow{-2}{*}{15} & \multirow{-2}{*}{0.52} & \multirow{-2}{*}{10/0} & $\gamma_{15}$: 1.7 & \multirow{-2}{*}{0.56} & \multirow{-2}{*}{10/0} & $\gamma_{15}$: 2.8 & \multirow{-2}{*}{0.58} & \multirow{-2}{*}{4/0} & $\gamma_{15}$: 9.5 \\
  \hline
\end{tabular} 
}}
\end{table}

As in Gazebo, a larger prediction horizon reduces the maximum deviation: $h=1$ violates the safety requirement for all configurations, $h=10$ does so for $\gamma_4 + \gamma_{15}$ at $\kappa_{\text{ref}} \in \{0.2, 0.9\}$, and $h=15$ keeps the maximum below \SI{0.69}{\meter} across all configurations and curvatures.
The curvature itself has only a mild effect on the maximum deviation but increases the time spent in \ac{ha} mode, as recovery into the contraction ellipsoid is more challenging at higher curvatures.

\cref{fig:circular_real_world_decision_module} reveals that the safety violations for $\gamma_4 + \gamma_{15}$ ($h=10$) occur when switching directly to $\gamma_{15}$ during an active \ac{ha} phase.
This also occurred for $\kappa_{\text{ref}}=0.2$, which additionally saw more frequent \ac{ha} activations.
On a side note, the brief \ac{ha} activation near \SI{70}{\sec} is a single-period false positive caused by a curvature discontinuity from ROS2 path replanning.
While this confirms that Gazebo simulations closely approximate real-world behavior, physical deployment still requires conservative safety margins since only $h=15$ maintained safety in all configurations.

To guarantee real-time schedulability on physical hardware, the Simplex decision module must introduce minimal computational overhead. 
Across all circular real-world experiments (\num{81749} decision cycles), the average execution latency is \SI{0.0281}{\milli\second} (standard deviation: \SI{0.0396}{\milli\second}), with a worst-case latency of \SI{4.1303}{\milli\second}. 
Compared to the system's \SI{50}{\milli\second} control period, these overheads are negligible and take less than \num{0.1}\% on average and under \num{10}\% in the worst case. 
Though worst-case latencies require deeper analysis, these results validate the feasibility of our context-dependent runtime \ac{roa} with minimal overhead on resource-constrained embedded platforms.

\subsubsection{Real-World Scenario}
\label{subsubsec:high_performance_controller}

\begin{table}
    \centering
    \caption{Simplex metrics under BC ($h=12$), skipping first/last blocks.}
    \label{tab:ha_real_urz_h12_bc_skipped}
    {\footnotesize
\setlength{\tabcolsep}{6pt}
\begin{tabular}{@{}cc@{}}
  \toprule
  $\max(T_{\text{HA}})$ [s] & $\max(|e_y|)$ [m] \\
  \midrule
  0.850 & 0.3741 \\
  1.050 & 0.3877 \\
  0.751 & 0.4109 \\
  \bottomrule
\end{tabular}
}

\end{table}
 
To evaluate a realistic trajectory-following scenario, we applied the Simplex controller to an urban test track. 
This time, the \ac{hp} controller is an \ac{nn} trained using behavioral cloning (BC) via the CORL library~\mycite{tarasov2022corl} for \num{50} epochs on offline expert data acquired through manually controlling the prototype.
We configured the Simplex framework with $h=12$, as the previous section showed that $h=10$ can violate safety constraints.

The trajectory and corresponding metrics for three runs under configuration $\gamma_4 + \gamma_{15}$ are shown in \cref{fig:real_world_traj} and \cref{tab:ha_real_urz_h12_bc_skipped}. 
The Simplex framework guarantees physical safety by keeping the maximum lateral deviation $\max(|e_y|)$ below the \SI{1.0}{\meter} threshold. 
Moreover, the \ac{hp} controller required only one brief intervention per run on average ($N_{\text{HA}}=3$, $\max(T_{\text{HA}}) = \SI{1.050}{\second}$), showing that our less conservative \ac{roa} successfully expanded the operating envelope.
This is visually evident in \cref{fig:real_world_traj}: the nominal path is smooth and continuous (blue), with the \ac{ha} controllers successfully executing only at the final critical curve (orange and purple).
\ac{ha} activations at start and end positions are due to discontinuities and are ignored in this analysis.

\section{Conclusions}
\label{sec:conclusions}

Runtime assurance (\ac{rta}) frameworks certify or correct \ac{hp} control actions proposed by uncertified controllers to maintain safety, but face a central trade-off: runtime reachability minimizes conservatism by using current context but imposes computational overhead on resource-constrained devices, whereas design-time global \ac{roa} approximations enable minimal runtime overhead at the cost of high conservatism.

To address this, we identify stability and safety artifacts of a controller at design-time to enable low-overhead, context-dependent runtime \ac{roa} construction. Our two-stage procedure combines contraction analysis with Taylor-model reachability, using parameter-space partitioning and binning to curb wrapping effects. This yields the controller's \textit{recovery profile} $\mathcal{R}$, which forms the basis for our multi-candidate Simplex architecture. By supporting both \ac{hp}-to-\ac{ha} and \ac{ha}-to-\ac{ha} switching, this architecture leverages complementary \ac{roa} coverage of the \ac{ha} candidates to enlarge the system's safe operating envelope.

We evaluated the framework on robotic path tracking using \ac{ode} simulations, Gazebo, and a physical differential-drive robot. 
While the simulations confirmed that the theoretical approach maintains safety, physical deployment revealed that unmodeled dynamics require adjusting the prediction horizon to introduce safety margins. Furthermore, a preliminary latency analysis verified that the approach incurs a minimal runtime overhead, averaging just \SI{0.0281}{\milli\second} or under \num{0.1}\% of the \SI{50}{\milli\second} control period on average but simultaneously motivates further investigation of the discrepancy with the observed worst-case latency of \SI{4.13}{\milli\second}.

Along with that, several promising directions remain for future work. 
First, a formal timing and schedulability analysis is warranted to investigate the gap between average and worst-case execution latencies. 
Second, the necessity of Lyapunov-based hysteresis for \ac{ha}-to-\ac{hp} transitions should be evaluated, as the large certified \ac{roa} boundary may naturally prevent chattering.
Third, incorporating richer physical error dynamics—such as wheel slip, actuator latency, and acceleration bounds—directly into the offline verification pipeline could replace heuristic prediction horizon tuning with fully certified, theoretically grounded bounds. 
Finally, assessing the framework's generalizability to other system classes and higher-dimensional state spaces should be targeted.

\section*{Acknowledgment}
This work was supported by the German Academic Exchange Service (DAAD) with funds from the German Federal Ministry of Education and Research (BMBF) within the Programme for Project-Related Personal Exchange (PPP) with Portugal, project RobustSAR (project ID 57764655).
This work was also supported by the Funda\c{c}\~{a}o para a Ci\^{e}ncia e a Tecnologia (FCT) under the RobustSAR project (ref.~2024.07838.CBM) and the LASIGE Research Unit (ref.~UID/00408/2025, DOI: \url{https://doi.org/10.54499/UID/00408/2025}).
We thank Stefan Schulze, M.\,Sc., for helping us prepare the behavioral-cloning controller.

\bibliographystyle{IEEEtran}
\bibliography{bibliography.bib}

\end{document}